\documentclass[sn-mathphys-num]{sn-jnl}

\usepackage{graphicx}%
\usepackage{multirow}%
\usepackage{amsmath,amssymb,amsfonts}%
\usepackage{amsthm}%
\usepackage{mathrsfs}%
\usepackage[title]{appendix}%
\usepackage{xcolor}%
\usepackage{textcomp}%
\usepackage{manyfoot}%
\usepackage{booktabs}%
\usepackage{algorithm}%
\usepackage{algorithmicx}%
\usepackage{algpseudocode}%
\usepackage{listings}%

\usepackage{longtable}
\usepackage{tabularx}
\usepackage{multirow}
\usepackage{float}
\usepackage{color}

\usepackage{multirow}
\usepackage{multicol}
\usepackage{subcaption}

\theoremstyle{thmstyleone}%
\theoremstyle{thmstyletwo}%

\theoremstyle{thmstylethree}%

\begin{document}

\title[Article Title]{Analysis of the Two-Step Heterogeneous Transfer Learning for Laryngeal Blood Vessel Classification: Issue and Improvement}


\author[1]{\fnm{Xinyi} \sur{Fang}}\email{xinyi.fang@mpu.edu.mo}

\author[1]{\fnm{Xu} \sur{Yang}}\email{xuyang@mpu.edu.mo}

\author[1]{\fnm{Chak Fong} \sur{Chong}}\email{chakfong.chong@mpu.edu.mo}

\author[1,2]{\fnm{Kei Long} \sur{Wong}}\email{keilong.wong@mpu.edu.mo}

\author*[1]{\fnm{Yapeng} \sur{Wang}}\email{yapengwang@mpu.edu.mo}

\author[3]{\fnm{Tiankui} \sur{Zhang}}\email{zhangtiankui@bupt.edu.cn}

\author[4]{\fnm{Sio-Kei} \sur{Im}}\email{marcusim@ipm.edu.mo}

\affil*[1]{\orgdiv{Faculty of Applied Sciences}, \orgname{Macao Polytechnic University}, \orgaddress{\city{Macao}, \country{China}}}

\affil[2]{\orgdiv{Department of Computer Science and Engineering}, \orgname{University of Bologna}, \orgaddress{\city{Bologna}, \country{Italy}}}

\affil[3]{\orgdiv{School of Information and Communication Engineering}, \orgname{Beijing University of Posts and
Telecommunications
}, \orgaddress{\city{Beijing}, \country{China}}}

\affil[4]{\orgname{Macao Polytechnic University}, \orgaddress{\city{Macao}, \country{China}}}


\abstract{Accurate classification of laryngeal vascular as benign or malignant is crucial for early detection of laryngeal cancer. However, organizations with limited access to laryngeal vascular images face challenges due to the lack of large and homogeneous public datasets for effective learning. Distinguished from the most familiar works, which directly transfer the ImageNet pre-trained models to the target domain for fine-tuning, this work pioneers exploring two-step heterogeneous transfer learning (THTL) for laryngeal lesion classification with nine deep-learning models, utilizing the diabetic retinopathy color fundus images, semantically non-identical yet vascular images, as the intermediate domain. Attention visualization technique, Layer Class Activate Map (LayerCAM), reveals a novel finding that yet the intermediate and the target domain both reflect vascular structure to a certain extent, the prevalent radial vascular pattern in the intermediate domain prevents learning the features of twisted and tangled vessels that distinguish the malignant class in the target domain, summarizes a vital rule for laryngeal lesion classification using THTL. To address this, we introduce an enhanced fine-tuning strategy in THTL called Step-Wise Fine-Tuning (SWFT) and apply it to the ResNet models. SWFT progressively refines model performance by accumulating fine-tuning layers from back to front, guided by the visualization results of LayerCAM. Comparison with the original THTL approach shows significant improvements. For ResNet18, the accuracy and malignant recall increases by 26.1\% and 79.8\%, respectively, while for ResNet50, these indicators improve by 20.4\% and 62.2\%, respectively.}




\maketitle

\section{Introduction}\label{Introduction}

Laryngeal cancer is one of the most frequently reported cancers in the head and neck. According to \citep{deng2020global}, by 2017, instances of laryngeal cancer have increased by $58.6\%$ in 27 years worldwide. At the same time, as the mortality rate increases, the average age of the disease slips. Although the final diagnosis of laryngeal lesions relies on clinical biopsy results, different types and stages of lesions can be reflected in the vascular structure changes of the larynx \citep{lukes2013narrow}.
In otolaryngology endoscopy, the emerging narrow-band image (NBI) technique is favored over traditional white light imaging in the diagnosis and early detection of lesions \citep{chabrillac2021narrow}  since it keeps the blue and the green light with specific wavelengths, which can better show the morphology of the mucosal epithelium and the epithelial vascular. With the development of deep learning, combining deep learning techniques and NBI for computer-aided diagnosis, it is possible to differentiate different kinds of lesions and obtain results that are extremely close to pathological diagnosis \citep{he2021deep}.

Transfer learning is a promising technique widely used to improve performance in medical image classification tasks, especially when the scale of the medical image dataset is normally small. Transfer learning refers to inheriting the knowledge acquired from the source domain to improve the performance of corresponding tasks in the target domain, such as the classification task. Applying transfer learning in image analysis is ubiquitous by combining the convolutional neural networks (CNNs) and reusing partial model parameters pre-trained on the source domain, commonly a larger dataset scale, such as ImageNet \citep{Oquab}. 

Transfer learning in medical images is mainly heterogeneous, meaning the source and target domains do not share the same type of images. Currently, many transfer learning in medical image classification is one-step heterogeneous transfer learning, in which the knowledge learned from the source domaine, e.g., the model's parameters trained in the source domain, will be used directly to further fine-tune in the target domain, a medical image dataset.

However, \citet{tan2017distant} point out that the performance of one-step heterogeneous transfer learning would be lower than expected if the features between the source domain and target domain are highly distinct. \citet{Matsoukas_2022_CVPR} prove that the features from ImageNet have little reuse on the chest x-ray and lymph node-stained sections. \citet{raghu2019transfusion} also conclude that ImageNet as the source domain does not significantly improve the performance of diabetic retina diagnosis and chest x-ray detection. 

One of the solutions to minimize the effects of ImageNet is the two-step heterogeneous transfer learning, which adds an intermediate domain to empower the model with more similar features to apply in the target domain. Currently, studies about the two-step heterogeneous transfer learning remain rare, especially in selecting the appropriate intermediate domain.  

Existing studies show that successful two-step transfer learning could rely on selecting an appropriate intermediate domain with features similar to the target domain dataset. However, for medical image kinds such as laryngeal blood vessels, there is a lack of homogeneous, large-scale public datasets to serve as intermediate domains to provide the model with similar features for the task of laryngeal blood vessel classification in the target domain. Thus, it is crucial to explore the feasibility of using another organ's dataset as the intermediate domain, e.g., another type of blood vessel, for the laryngeal blood vessel classification task.

Therefore, two hypotheses are waiting for us to answer in this work. \textit{\textbf{Hypothesis (1)}}: performing two-step heterogeneous transfer learning in laryngeal blood vessel classification by transferring the knowledge of blood vessels from other organs as the intermediate domain is feasible. \textit{\textbf{Hypothesis (2)}}: using ImageNet as the source domain in laryngeal blood vessel classification is a good choice.

The main contributions of this article are summarized as the following:
\begin{itemize}
\item This is the first work to investigate the effectiveness of two-step heterogeneous transfer learning on laryngeal blood vessel binary classification task by comparing the classification performance of nine deep learning models. The source domain is ImageNet, the intermediate domain is the color fundus photographies of Diabetic Retinopathy \citep{diabetic-retinopathy-detection}, and the target domain is the vocal fold subepithelial blood vessels images \citep{CENBIdataset}. Experiments demonstrate that the overall performance drops significantly when only fine-tuned the last layer in the second step of THTL (accuracy drops 12.2\% on average for nine models, and recall of the malignant class drops 42.9\%).
\item We use an attention mechanism called LayerCAM \citep{layerCAM} combined with ResNet18 to visualize the area of interest for each Layer of the model and explore the key features used for image classification. The visualization reveals a novel finding that the color fundus vascular pattern shows a radial pattern, missing the condition for judging malignancy of the laryngeal vessels (a twisted vascular pattern), even if they share the characteristic of being blood vessels.
\item Moreover, to address the performance drop caused by using the Diabetic Retinopathy images as the intermediate domain in THTL, with only the last layer fine-tuned, we propose a fine-tuning strategy called Step-Wise Fine-Tuning (SWFT) to gradually increase the number of layers fine-tuned from back to front for the model, resulting a substantial improvement. ResNet18 increases 26.1\% for accuracy, as well as ResNet50 increases 20.4\%. Such improvement in results shows that the proposed SWFT can overcome the performance drop caused by the inappropriate intermediate domain. However, compared to one-step transfer learning with all layers updated, SWFT achieves comparable accuracy, albeit a little less. Notably, the shallow model may need more steps to be fine-tuned in SWFT.
\item We also obtain that there is no significant improvement in performance by using ImageNet as a source domain than training from scratch, while the total computational time is shortened.
\item We summarize the best-performing models in four laryngeal blood vessel classification scenarios. For models pre-trained on ImageNet and then fine-tune the final layer in the target domain,  ViT-B/16 performs the best in one-step heterogeneous transfer learning. VGG19 performs the best in two-step heterogeneous transfer learning. For models pre-trained on ImageNet and then fully trained on the laryngeal vessels dataset, the classification accuracy of EfficientNetV2-S is the highest. As for models training from scratch on the laryngeal vessels dataset, the InceptionV3 performs better overall.
\end{itemize}

\section{Related works}\label{sec2}
\subsection{Definition of Transfer Learning}
A formal definition is given by \citet{pan2010survey}, a source domain $\mathcal{D}_S= \{\mathcal{X}_S, P(X_S)\}$ and a target domain $\mathcal{D}_T = \{\mathcal{X}_T, P(X_T)\}$, where the $\mathcal{X}$ and $P(X)$ are the feature space and marginal data distribution of the corresponding domain. Given a source domain learning task $\mathcal{T}_S$ and a target domain learning task $\mathcal{T}_T$, where the task corresponding to each domain contains that domain's label space $\mathcal{Y}$ and predictive function $f(\cdot)$. Using knowledge learned from the $\mathcal{D}_S$ and $\mathcal{T}_S$, transfer learning aims to improve the performance of the target predictive function $f_T(\cdot)$ in the $\mathcal{D}_T$. Notice that neither domains nor tasks are the same. 

Moreover, transfer learning can be categorized into homogeneous and heterogeneous based on the difference between the source and target domain. Homogeneous transfer learning strictly restricts the feature space, and the label space of the source and target domain must be the same. Failing to meet any of these conditions is a heterogeneous transfer learning \citep{day2017survey}.

In this work, we propose to investigate the effectiveness of two-step heterogeneous transfer learning for laryngeal blood vessel classification by given the definition of the intermediate domain $\mathcal{D}_I= \{\mathcal{X}_I, P(X_I)\}$, the learning task of the intermediate domain $\mathcal{T}_I = \{\mathcal{Y}_I, f_I(\cdot)\}$, and a strict condition, which is $\mathcal{X}_S \neq \mathcal{X}_I \neq \mathcal{X}_T$, $\mathcal{Y}_S \neq \mathcal{Y}_I \neq \mathcal{Y}_T$, and $\mathcal{T}_S \neq \mathcal{T}_I \neq \mathcal{T}_T$.

\subsection{One-step Heterogeneous Transfer Learning}
In medical image classification tasks, directly transferring the models' parameters from ImageNet to the target domain is a conventional approach and usually obtains a well-performed result, such as some works of laryngeal lesions classification \citep{araujo2019learned, xiong2019computer}. However, they lack of verifying the effectiveness of using ImageNet as the source domain for transfer learning. 

Using ImageNet as the source domain might not always promising a good results.  \citet{heker2020joint} points out that in some circumstances, under medical image classification tasks, using ImageNet as the source domain might not bring an expected performance. Still, there is a chasm between the medical and natural images. \citet{alzubaidi2020optimizing} prove that the breast cancer classification task will perform better if the source domain transferred is closed to the target domain, not ImageNet. \citet{Xie_2018_ECCV_Workshops} obtain that for chest x-ray classification task, models pre-trained on gray-scale ImageNet exceeds those pre-trained on original ImageNet.

To the best of the authors' knowledge, it is an open question whether using ImageNet as a source domain for classifying laryngeal blood vessels is effective. 

\subsection{Two-Step Heterogeneous Transfer Learning}
The idea of sequential transfer learning \citep{sequential_transfer_learning} in natural language processing is utilized in medical image classification, as manifested in knowledge can be transferred from the source domain to the intermediate domain first, then transferring the knowledge learned from the intermediate domain to the target domain.  

\citet{de2019double} obtain improved results in breast cancer histopathologic image classification. They use ImageNet as the source domain to get the pre-trained models, then train a model in an intermediate domain with histopathologic images to acquire the knowledge to exclude the blank space of a breast tumor histopathologic image in the target domain. \citet{alkhaleefah2020double} compare the classification performance of one-step and two-step heterogeneous transfer learning in mammogram images and conclude that two-step heterogeneous transfer learning outperforms one-step’s. Notice that the source domain is ImageNet, and the dataset in the intermediate and target domains are mammogram images. The paper of \citet{alzubaidi2021novel} demonstrates that the classification performance of diabetic foot ulcers is enhanced by transferring the knowledge learned from the skin cancer to the feet' skin image first, then transferring the knowledge to the diabetic foot ulcers. \citet{meng2022tl} apply two-step heterogeneous transfer learning to detect COVID-19. They use ImageNet as the source domain, then use the tuberculosis (TB) CT image dataset as the intermediate domain for the model to learn the features close to the target domain. The target domain is COVID-19 CT images. This two-step heterogeneous transfer learning achieves a recognition accuracy of 93.24\%.

Despite the entities above being successful, there are limitations to two-step heterogeneous transfer learning. Whether in terms of imaging modality or semantics, the datasets in the intermediate domains for those studies are very close to those in the target domains. This is feasible for those high-intensity discussed medical image types. However, it is challenging to find a publicly available large-scale dataset as the intermediate domain for two-step heterogeneous transfer learning for those relatively less discussed medical image types, such as laryngeal blood vessels. Therefore, it is critical to seek an alternative intermediate domain that might be formed by different imaging modalities and semantically nonidentical to the laryngeal blood vessel. 

\subsection{Deep-Learning Models Applied in Medical Image Classification}
In this work, different types of models will be employed to verify the feasibility of two hypotheses we proposed. \citet{morid2021scoping} statistics the frequently used deep learning models in medical image analysis from 2012 to 2020. In particular, Inception \citep{Szegedy_2015_CVPR}, ResNet \citep{he2016deep}, and Visual Geometry Group (VGG) \citep{simonyan2014very} are commonly used in endoscopic images. We select Inception\-V3 \citep{Szegedy_2016_CVPR}, ResNet18, ResNet50, and VGG19 in our experiments. We also include DenseNet \citep{huang2017densely}, which is commonly practiced in lung studies, into consideration. The selection for DenseNet is DenseNet121 and DenseNet169. Additionally, the lightweight model MobileNet V2 \citep{Sandler_2018_CVPR} is efficient and performs excellently in soft tissue classification \citep{Arfan_mobilenet}. EfficientNet\-B0 has made its debut in laryngeal disease classification \citep{cho2021diagnostic}, with the highest accuracy and relatively low memory consumption, proving the value of EfficientNet in the classification of laryngeal diseases. For laryngeal image classification using another imaging technique called probe-based confocal laser endomicroscopy (pCLE), Vision Transformer (ViT) \citep{dosovitskiy2021an} shows a good performance \citep{Transformer_Cao}. Thus, we also conclude MobileNet V2, and two state-of-art models (the EfficientNetV2-S from EfficientNetV2 \citep{pmlr-v139-tan21a} and ViT-B/16 from Vision Transformer) in our experiments.

\subsection{Motivation of this work}
Although the current dominant choice of source domain for medical image classification with transfer learning is ImageNet, some studies state that using ImageNet as the source domain brings unsatisfactory performance instead. Therefore, using two-step heterogeneous transfer learning, empowering the model after training in the intermediate domain is necessary for some medical image classification tasks.  

However, current studies about the two-step heterogeneous transfer learning remain rare, especially for selecting the appropriate intermediate domain or summarizing rules for classifying laryngeal blood vessels. Moreover, considering the data scarcity of this kind of medical image, it becomes more challenging to choose the semantically identical intermediate domain for two-step transfer learning. Therefore, it is critical to investigate the semantically nonidentical intermediate domain.  

Additionally, verifying the effectiveness of using ImageNet as the source domain for one-step heterogeneous transfer learning in laryngeal blood vessel classification is necessary for identifying the importance of performing a two-step transfer learning strategy.

\section{Methodology}\label{Methodology}
In this section, three public datasets, and the data preparation and pre-processing work, will be introduced. Also, the proposed Step-Wise Fine-Tuning method will be introduced, as well as the evaluation metrics for classification performance.

\subsection{Datasets and Data Pre-processing Methods}\label{Sec: Datasets and Data Pre-processing Methods}
\subsubsection{ImageNet (Source Domain)}\label{ImageNet Dataset}
In this paper, ImageNet is used as the source domain task for experiments. ImageNet is the most frequently used, well-known dataset for pre-trained models in the image field, the ImageNet Large Scale Visual Recognition Challenge 2012 (ILSVRC2012) of ImageNet \citep{ImageNet}. This dataset contains 1000 categories of natural images over one million. The weights of pre-trained models based on ImageNet are used as the initial weights for later training.

\subsubsection{CE-NBI Dataset (Target Domain)}\label{CE-NBI Dataset}

\begin{figure}[b]
\centering
\includegraphics[width=\linewidth]{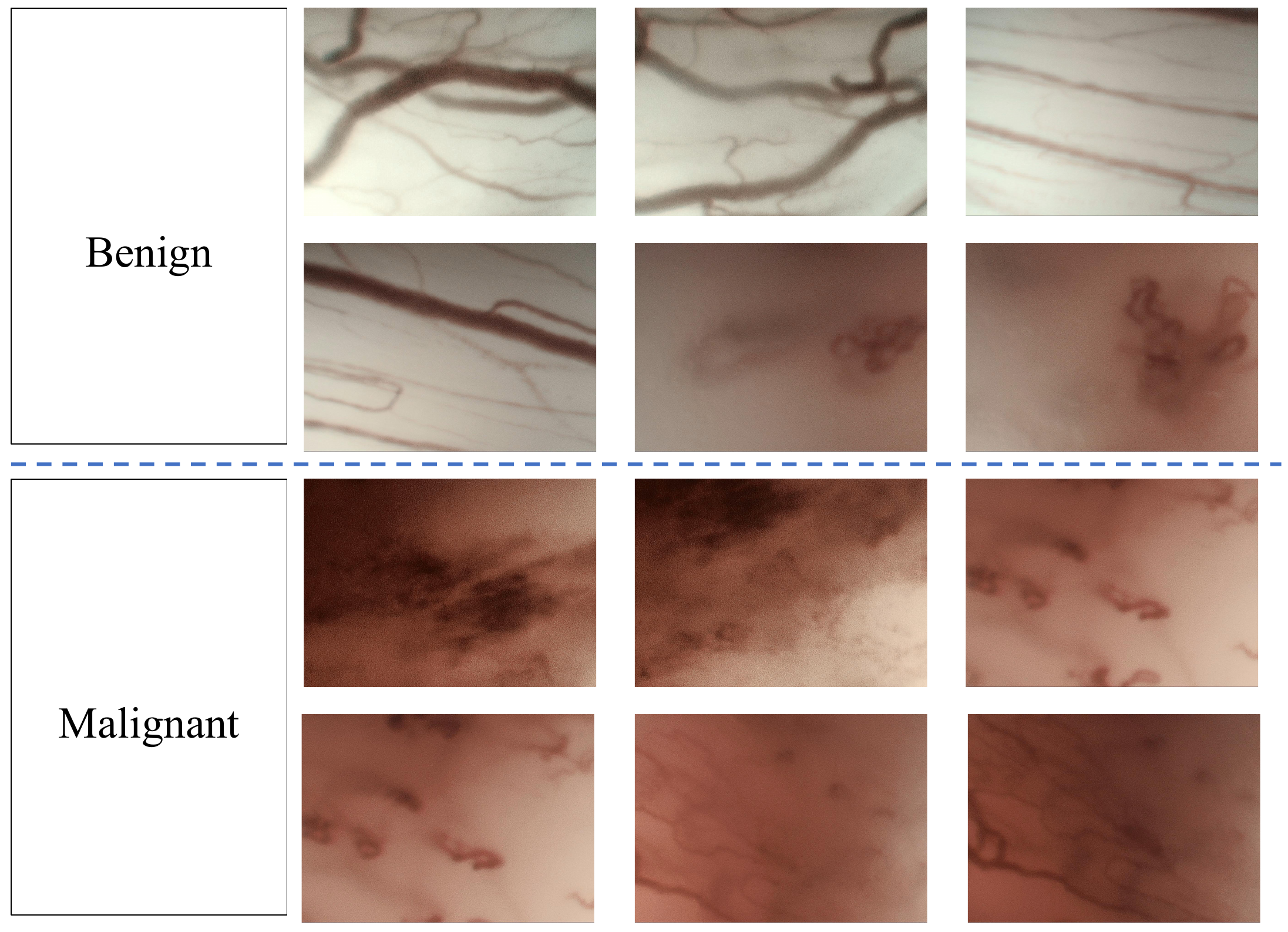}
\caption{Samples of CE-NBI dataset.} \label{fig: CE-NBI samples}
\end{figure}

The Contact Endoscopy combined with Narrow Band Imaging (CE-NBI) is released by \citet{CENBIdataset} in 2019. This dataset is used as the target domain task for laryngeal blood vessel classification. In this dataset, the subepithelial blood vessels of the vocal folds are enhanced and magnified. The dataset contains 11144 images of vocal fold subepithelial blood vessels, categorized into benign and malignant according to the type of lesions. Depending on laryngeal histopathology diagnostics, both benign and malignant are further subdivided into different lesions. The benign category includes Amyloidosis, Cyst, Granuloma, and other eight kinds of diseases; the malignant are subdivided into squamous cell carcinoma (SCC), high grade dysplasia, and Carcinoma in situ. As the target domain task of our research, to better avoid the class imbalance issue, we use the lesion type label to classify the CE-NBI images into benign and malignant classes. The samples of this dataset are shown below in Fig. \ref{fig: CE-NBI samples}. The first two rows are the samples of the benign class, and the last two are the malignant class samples.

In this dataset. the images are randomly divided as the ratio of $7:2:1$ into training, validation, and testing set, using random seed 123. Thus, there are 7803 images for training, 2228 for validation, and 1113 for testing. Detailed data volume statistics can be found in \autoref{tab: data volume CE-NBI}.

\begin{table}[h] 
\caption{Data Volume Statistics of CE-NBI Dataset\label{tab: data volume CE-NBI}}
\begin{tabular}{@{}lcccc@{}}
\toprule
\textbf{Class}	& \textbf{Training Set}	& \textbf{Validation Set} & \textbf{Testing Set} & \textbf{Total}\\
\midrule
Benign		& 5361  & 1531 & 765 & 7657\\
Malignant	& 2442	& 697 & 348 & 3487 \\
\bottomrule
\end{tabular}
\end{table}

Also, we apply data augmentation techniques during the training phase by resizing the images to meet the model's requirement and randomly horizontally flipping the images. For the validation and testing phases, we resize the images only. Moreover, image normalization is applied in all three phases, using $mean = [0.485, 0.456, 0.406]$ and standard deviation equals $[0.229, 0.224, 0.225] $.

\subsubsection{DR Dataset (Intermediate Domain)}\label{DR Dataset}

\begin{figure}[b]
\centering
\includegraphics[width=\linewidth]{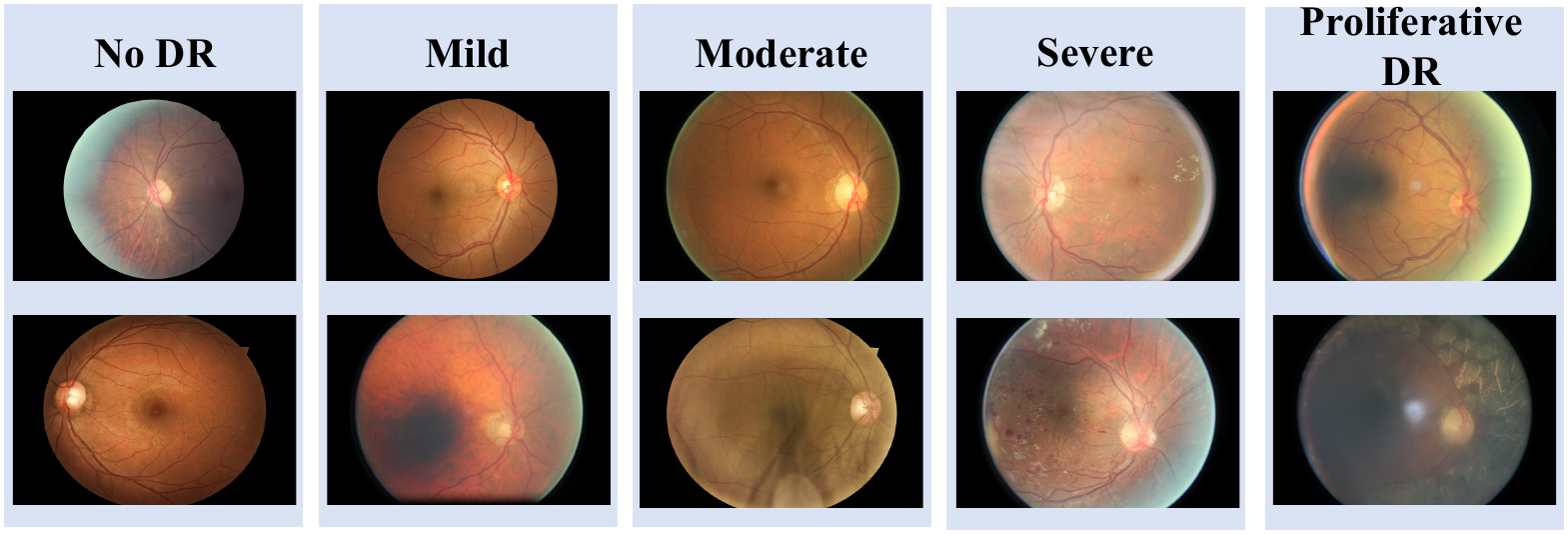}
\caption{Samples of DR dataset.} \label{fig: DR samples}
\end{figure} 

We aim to utilize a modality and semantically nonidentical dataset to the target domain dataset, but only if under the premise that they are both blood vessels. A dataset of Diabetic Retinopathy (DR) Detection publishes on Kaggle \citep{diabetic-retinopathy-detection} is used as the intermediate dataset. The modality of this dataset is digital color fundus photography, which is different from the CE-NBI endoscopy image. This dataset serves to identify the severity of DR by color fundus photographies into five classes: no sign of Diabetic Retinopathy, a mild symptom of DR, a moderate symptom of DR, a severe symptom of DR, and Proliferative DR. The task is also different to the CE-NBI dataset. Therefore, the DR dataset satisfies the restrictions of the intermediate domain for two-step heterogeneous transfer learning for blood vessel classification.

\begin{table}[h] 
\caption{Data Volume Statistics of Diabetic Retinopathy Detection Dataset\label{tab: data volume DR}}
\begin{tabular}{@{}lccc@{}}
\toprule
\textbf{Class}	& \textbf{Training Set}	& \textbf{Validation Set} & \textbf{Total}\\
\midrule
No DR		& 18067  & 7743 & 25810\\
Mild	& 1710	& 733 & 2443 \\
Moderate	& 3704	& 1588 & 5292 \\
Severe	& 611	& 262 & 873 \\
Proliferative DR	& 495 & 213 & 708 \\
\bottomrule
\end{tabular}
\end{table}

We only use the training set of this public dataset since the competition is closed and released test dataset are not labelled. Therefore, we treat the training set as the whole dataset (DR dataset), and further split it into training and validation set for our experiment. We use the training set to train the models and select the model with the highest performance on the validation set. Thus, the testing set is optional. The DR dataset contains 35126 images, randomly divided into a training and a validation set by the ratio of $7:3$, with the random seed 123. Thus, the training set contains 24587 images, and the validation set contains 10539 images. Samples of the DR dataset can be found in Fig. \ref{fig: DR samples}. Two samples of the current class are in each column.  \autoref{tab: data volume DR} shows each class's detailed data volume statistics. 

The data augmentation configuration follows the same operations as the CE-NBI dataset.

\subsection{Step-Wise Fine-Tuning Method}\label{sec:SWFT method}

\begin{figure}[b]
\centering
\includegraphics[width=\linewidth]{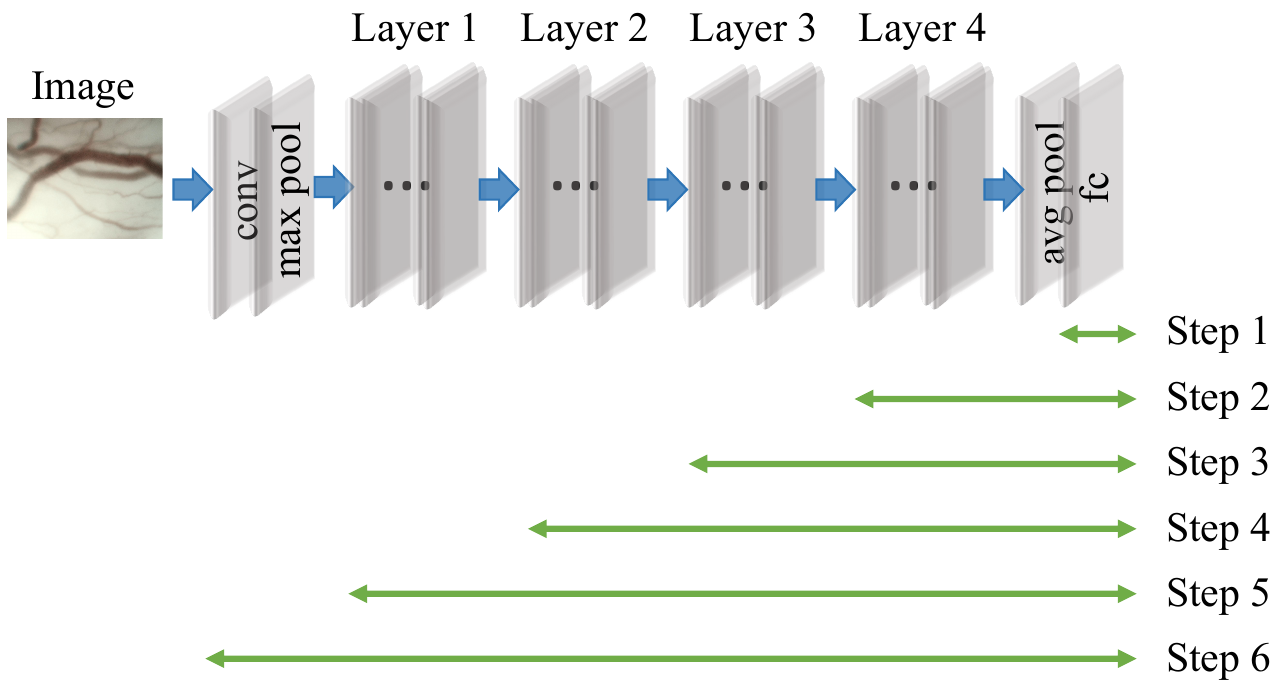}
\caption{Demonstration of Step-Wise Fine-Tuning for ResNet.} \label{fig: ResNet_SWFT}
\end{figure}

Since transfer learning is usually combined with deep-learning models or convolution neural networks (CNNs), several fine-tuning strategies have been adopted for medical image classification. As summarized in \citet{kandel2020transfer}, there are three common fine-tuning strategies for transfer learning: 
\begin{enumerate}
\item Remove and replace the original fully connected layer with a task-specific fully connected layer or a new classifier layer while keeping the entire network weights frozen; 
\item Remove the original fully connected layer, fine-tune the entire network weights with a small learning rate, and incorporating a new classifier layer tailored to the new task; 
\item Remove the original fully connected layer, only fine-tune the layers close to the output while keeping the layers close to input frozen, then tailor the new classifier to the new task. 
\end{enumerate}

Despite those common fine-tuning techniques summarized in \citet{kandel2020transfer}, some flexible fine-tuning techniques can also bring effectiveness improvements, such as differential evolution based fine-tuning \citep{Adaptive_Fine_Tuning} and block-wise fine-tuning in \citep{boumaraf2021new}, which they fine-tunes the last two residual blocks of ResNet18 while keeping other weights frozen. In addition, this work is inspired by layer-wise fine-tuning in \citet{tajbakhsh2016convolutional} and \citet{sharma2020effect}. They fine-tune the smallest unit in the CNN sequentially, such as the convolutional layer and the fully connected layer.

In this work, we introduce a different fine-tuning strategy called the Step-Wise Fine-Tuning strategy (SWFT) to gradually increase the number of layers fine-tuned from back to front for the model in the second step of two-step heterogeneous transfer learning. Our proposed methodology is not only distinguished from \citet{tajbakhsh2016convolutional} and \citet{sharma2020effect} in terms of the specific model and dataset employed, but we also incorporate the visualization of the LayerCAM as a reference for the number of steps in SWFT.

The examination and analysis of convolutional networks have demonstrated that lower layers, positioned closer to the input, primarily specialize in extracting texture-related features. Conversely, higher layers, situated closer to the output, are more class-specific and capture more semantic information, i.e., labels of the images \citep{zeiler2014visualizing, Ma_2015_ICCV}. Consequently, implementing our proposed SWFT facilitates the retention of intermediate domain knowledge within the lower-layer parameters of the model while simultaneously enabling the model's adaptation to the target domain in the top-layer parameters.  

The principle of SWFT is shown in Fig. \ref{fig: ResNet_SWFT}, with the ResNet model in the pytorch (ResNet has four major modules called Layer One to Four), the SWFT can be split into six steps. Step 1 fine-tunes the fully connected layer of the model; Step 2 fine-tunes Layer 4 and the fully connected layer; Step 3 fine-tunes Layer 3, Layer 4, and the fully connected layer. The layers involved in fine-tuning are added one by one up to Step 5. The last step, Step 6, is to update all parameters of the model. 

We have compared the performance of our proposed method with the conventional fine-tuning strategy, which involves fine-tuning either the last fully connected layer or all layers of the model. The experiment results show that our proposed SWFT achieves comparable accuracy to the regular fine-tuning approach. Particularly, the shallow model may need more Layers to be fine-tuned to achieve optimal performance.

\subsection{Evaluation Metrics}\label{Evaluation Metrics}

This article uses accuracy, precision, recall, F1-Score, and AUC as the evaluation metrics with the help of confusion matrix. The math expression of each metric is given below. The range of all matrices is from zero to the positive one.

Accuracy refers to the proportion of all corrected predicted images to all images, or we can express it as:
\begin{equation}
Accuracy = \frac{True\,Positives + True\,Negatives}{Total\,Number\,of\,Images}
\end{equation}

Precision refers to the proportion of actual positive-labeled images to the positive images that the model predicts.
\begin{equation}
Precision = \frac{True\,Positives}{True\,Positives + False\,Positives}
\end{equation}

Recall refers to the proportion of positive images the model predicts to the actual positive-labeled images.
\begin{equation}
Recall = \frac{True\,Positives}{True\,Positives + False\,Negative}
\end{equation}

F1-Score considers both precision and recall. Its value becomes large only if the precision and recall are both significant. When images are perfectly correctly classified, the F1-Score equals one.
\begin{equation}
F1\textnormal{-}Score = \frac{2*(Precision * Recall)}{Precision + Recall}
\end{equation}

AUC refers to measure the area under the receiver operating characteristic (ROC) curve. The performance of the model can be inferred from the value of AUC. The closer the value of AUC is to one, the better the model's classification performance is.

\section{Experiments Design}\label{sec: Experiments Design}
This section presents the the detailed process of experiments corresponding to the \textit{\textbf{hypothesis (1)}} and the \textit{\textbf{hypothesis (2)}}, including designs and configurations.

\subsection{Experiments for Hypothesis (1)}\label{Expt for H(1)}

\begin{figure*}[h]
\centering
\includegraphics[width=1\linewidth]{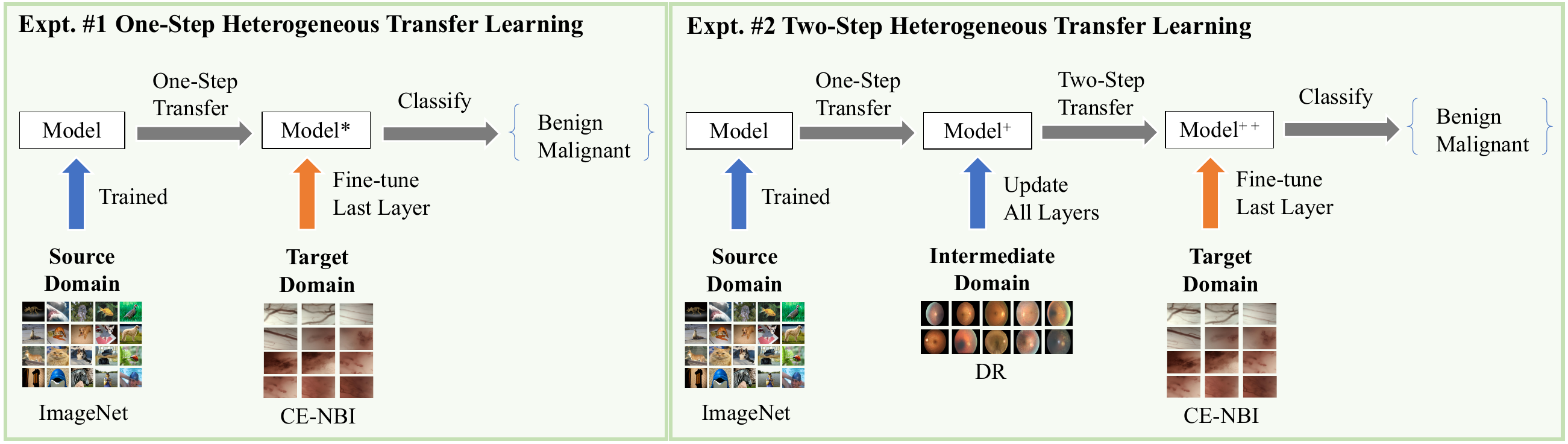}
\caption{Experiment Design for Hypothesis (1)}\label{fig: Experiment Design 1}
\end{figure*}

Expt.\#1 and Expt.\#2 aim to investigate the feasibility of two-step heterogeneous transfer learning in vocal fold subepithelial blood vessel classification tasks by comparing the performance difference between one-step and two-step heterogeneous transfer learning. The illustration is shown in Fig. \ref{fig: Experiment Design 1}.  

Expt.\#1 investigate the performance of one-step heterogeneous transfer learning. ImageNet is used as the source domain. CE-NBI dataset is used as the target domain. ImageNet pre-trained models, ResNet18, ResNet50, MobileNet V2, Inception\_V3, Densenet121, Densenet169, VGG19, ViT-B/16, and EfficientNetV2-S are used. The pre-trained weights are kept as the initial weights of models. 

To assess the impact of domain transfer on the target domain, we only fine-tune the fully connected layer of the model based on the target domain and modify the number of output classes to two (benign and malignant). The fine-tuned models are marked with the asterisk suffix $^*$. 

Expt.\#2 investigates the performance of two-step heterogeneous transfer learning. The ImageNet is also used as the source domain to transfer the pre-trained parameters of the models. Since we aim to capture the features of the intermediate domain in order to transfer it to the target domain, these models are fully trained and updated with all weights in the intermediate domain dataset, the DR dataset. Particularly, our objective extends beyond solely classifying the diabetic retina. We aim to transfer the knowledge of distinguishing blood vessels to the target domain task using the models trained in the intermediate domain. Consequently, we choose not to adjust the hyperparameters specifically for the models. To remain consistent, we utilize the hyperparameters listed in \autoref{sec: Details of Expt Config} for all experiments. These fully trained models are marked with the suffix $^+$. 

After that, we transfer model$^+$ and only fine-tune the model's last layer on the target domain to assess the impact of intermediate domain transfer on the target domain. In this phase, the fine-tuned models are marked with the suffix of $^{++}$, and then they are used to do the laryngeal blood vessel classification task.

\subsection{Experiments for Hypothesis (2)}\label{Expt for H(2)}
\begin{figure*}[h]
\centering
\includegraphics[width=1\linewidth]{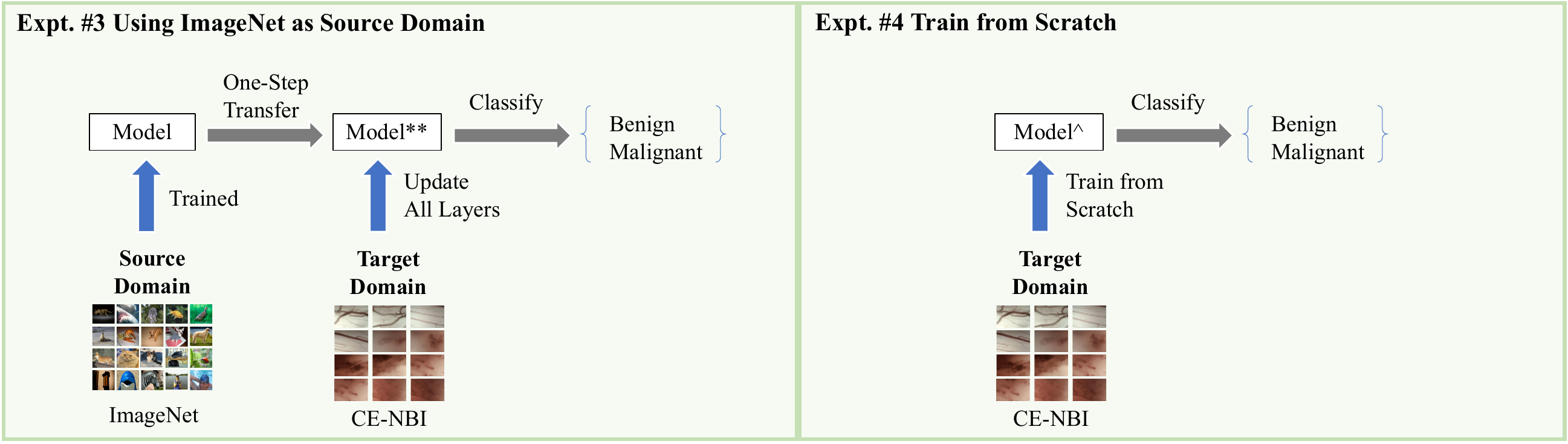}
\caption{Experiment Design for Hypothesis (2)}\label{fig: Experiment Design 2}
\end{figure*}

Expt.\#3 and Expt.\#4 aim to investigate the effectiveness of using ImageNet as the source domain in vocal fold subepithelial blood vessels classification by comparing the performance difference between models with or without pre-training on ImageNet. The designs are illustrated in Fig. \ref{fig: Experiment Design 2}. Models in Expt.\#3 are pre-trained on ImageNet and then updates all their parameters based on the target domain. These models are marked with the suffix $^{**}$. On the contrary, models in Expt.\#4 do not pre-trained on ImageNet. They all start with the random initial weights and only use the CE-NBI dataset for training, validating, and testing. They are marked with the suffix \textasciicircum. The results will be discussed in \autoref{sec: Results}. 

\subsection{Details of Experiment Configurations}\label{sec: Details of Expt Config}

The experiments are conducted using Python programming language, with all deep learning models developed using the PyTorch framework \citep{NEURIPS2019_bdbca288}. The NVIDIA RTX A6000 and NVIDIA GeForce RTX 2080 Ti GPUs are utilized  for the training, validation, and testing phases. We apply the following configurations to every experiment, including experiments \#1, \#2, \#3, and \#4. The batch size is set to $32$, with a maximum training epoch of $70$ and implementation of an early stopping technique. The parameters for the early stopping technique are set to a patience of $10$ and delta of $0$. The Adam optimizer \citep{adam} is utilized, with learning rate decay implemented through a learning rate step scheduler with $step\_size = 5$ and $gamma = 0.5$. The initial learning rate for EfficientNetV2-S and VGG19 models is set to $1e-4$, while for the other models it is set to $1e-3$. The loss function employed is Cross Entropy. The input size of the image varies depending on the model. For EfficientNetV2-S, the input size is adjusted to $384*384$. For InceptionV3, the input size is set to $299*299$. For other models, the input size is standardized at $224*224$. To avoid the contingency of the experiment, each model is run $3$ times and then the average of the results is recorded.

\section{Results}\label{sec: Results}
This section presents and discusses the results of experiments designed in \autoref{sec: Experiments Design}.

\subsection{Results to Verify the Hypothesis (1)}\label{Results for H(1)}

\begin{figure*}[h]
\centering
\subfloat[ROC Curve for Expt.\#1]{\includegraphics[width=0.5\linewidth]{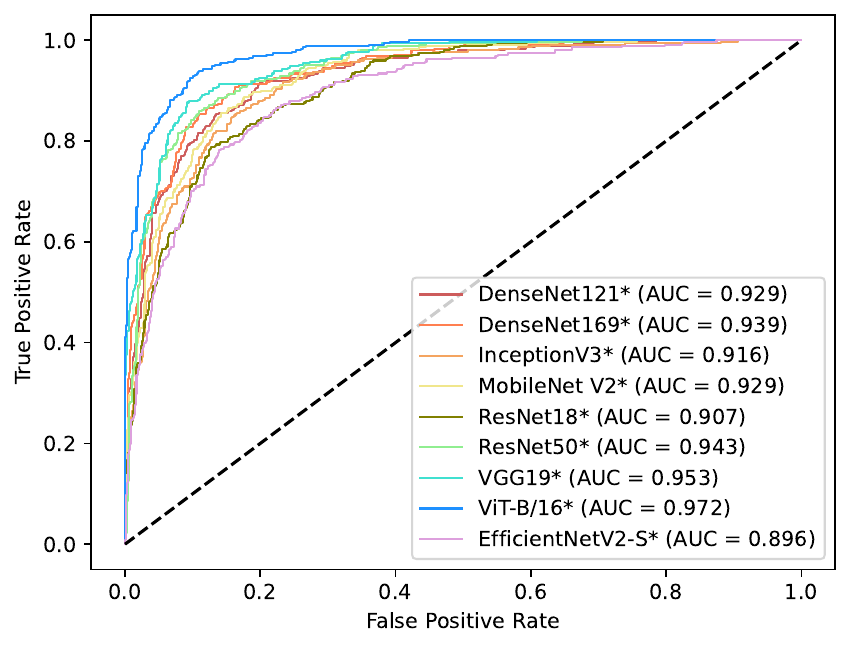}\label{subfig: ROC_Expt1}}
\hfil
\subfloat[ROC Curve for Expt.\#2]{\includegraphics[width=0.5\linewidth]{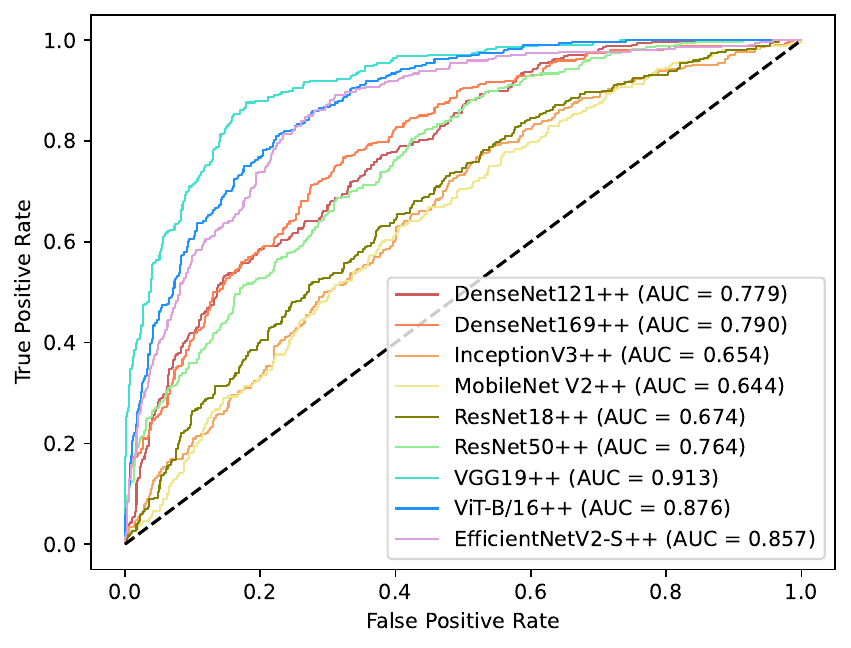}\label{subfig: ROC_Expt2}}
\\
\subfloat[ROC Curve for Expt.\#3]{\includegraphics[width=0.5\linewidth]{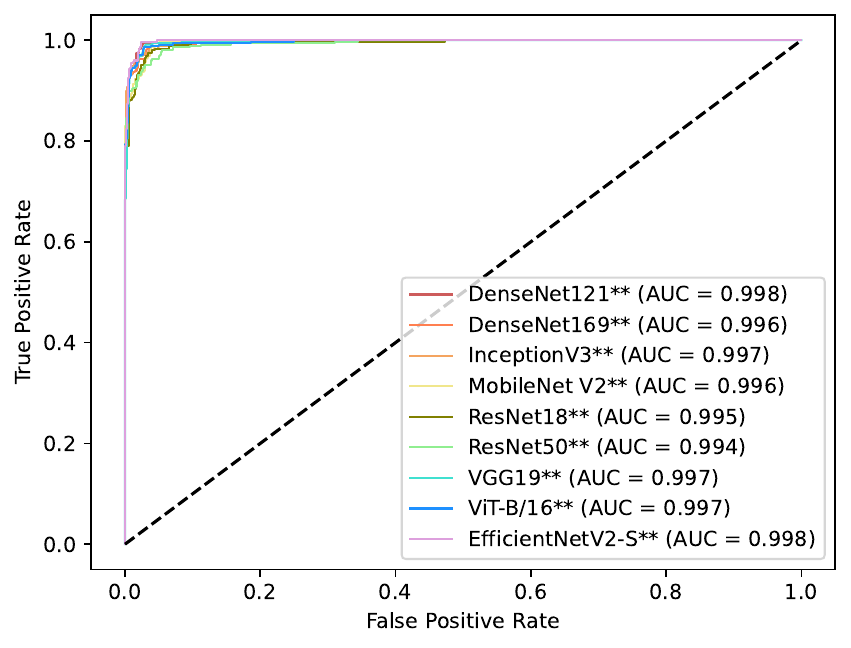}\label{subfig: ROC_Expt3}}
\subfloat[ROC Curve for Expt.\#4]{\includegraphics[width=0.5\linewidth]{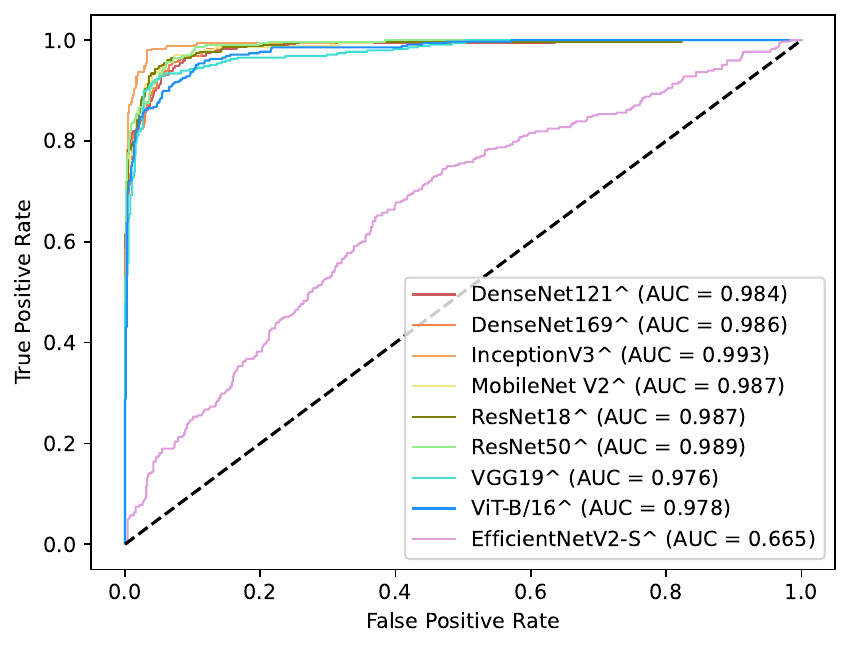}\label{subfig: ROC_Expt4}}
\caption{ROC Curves for Experiment One to Four}
\label{fig: ROC_Curves}
\end{figure*}

The results of Expt.\#1 and Expt.\#2 are presented in \autoref{tab: Expt1_and_Expt2_results}. In the table, every record is the average of three runs, the number of epochs is rounded up, and the rest of the data is kept to three decimal places. For both experiments, the highest result is marked as blue, while the lowest is marked as red for every column. The ROC curves for Expt.\#1 and Expt.\#2 are presented in the Fig. \ref{subfig: ROC_Expt1} and Fig. \ref{subfig: ROC_Expt2}. Given that we conduct three runs for each experiment, we select the run that is closest to the average testing accuracy presented in the \autoref{tab: Expt1_and_Expt2_results} in order to generate the ROC curve.

\begin{sidewaystable}
\caption{Results of Expt.\#1. and Expt.\#2.} \label{tab: Expt1_and_Expt2_results}
\begin{tabular*}{\textwidth}{@{\extracolsep\fill}llcccccccccc}
\toprule%
&&&& \multicolumn{2}{c}{Precision} & \multicolumn{2}{c}{Recall} & \multicolumn{2}{c}{F1-Score} &\\
\cmidrule{5-10}%
Expt. & Model & Epoch & Accuracy & B\footnotemark[1]{} & M\footnotemark[2]{} & B & M & B & M & AUC & Avg. Time\footnotemark[3]{}\\
\midrule
\multirow{9}{*}{\#1}& DenseNet121*                     & 42                              & 0.866                          & 0.883              & 0.823             & 0.929            & 0.730            & 0.905             & 0.773           & 0.929    & 58s                     \\ 
& DenseNet169*                     & 35                              & 0.880                          & 0.907              & 0.818             & 0.920            & 0.792            & 0.913             & 0.805   & 0.939  & 66s                       \\ 
& InceptionV3*                     & 32                              & 0.854                          & 0.873              & 0.805             & 0.922            & 0.706            & 0.897             & 0.752  & 0.915  & 70s                       \\ 
& MobileNet V2*                   & 44                              & 0.856                          & 0.878              & 0.799             & 0.918            & 0.719            & 0.897             & 0.757   & 0.928  &  48s                      \\ 
& ResNet18*                        & 42                              & 0.836                          & \textcolor{red}{0.858}              & 0.774             & 0.911            & \textcolor{red}{0.670}            & 0.884             & 0.718            & 0.908   & 46s                      \\
& ResNet50*                        & \textcolor{blue}{55}                              & 0.886                          & 0.906              & 0.837             & 0.930            & 0.788            & 0.918             & 0.812      & 0.943   & 56s                      \\ 
& VGG19*                          & 43                              & 0.878                          & 0.907              & 0.811             & 0.915            & 0.794            & 0.911             & 0.802     & 0.940  & 76s                       \\ 
& ViT-B/16*                      & 50                              & \textcolor{blue}{0.915}                          & \textcolor{blue}{0.930}              & \textcolor{blue}{0.879}             & \textcolor{blue}{0.947}            & \textcolor{blue}{0.844}            & \textcolor{blue}{0.939}             & \textcolor{blue}{0.861}          & \textcolor{blue}{0.972}    & 67s                     \\ 
& EfficientNetV2-S*             & \textcolor{red}{37}                              & \textcolor{red}{0.820}                          & 0.865              & \textcolor{red}{0.719}             & \textcolor{red}{0.875}           & 0.701            & \textcolor{red}{0.870}             & \textcolor{red}{0.710}           & \textcolor{red}{0.879}   & 109s                     \\\midrule
\multirow{9}{*}{\#2}& DenseNet121$^{++}$                     & 26                              & 0.750                          & 0.761              & 0.691             & 0.927            & 0.360            & 0.836             & 0.473   & 0.778  & 28s      \\ 
& DenseNet169$^{++}$                     & 31                              & 0.733                          & 0.734              & 0.724             & 0.959            & 0.236            & 0.832             & 0.355        & 0.789 & 34s       \\ 
& InceptionV3$^{++}$                     & 25                              & 0.689                          & \textcolor{red}{0.694}              & 0.544             & \textcolor{blue}{0.982}            & \textcolor{red}{0.046}            & 0.813             & \textcolor{red}{0.085}                & 0.657  & 30s      \\ 
& MobileNet V2$^{++}$                   & \textcolor{red}{23}                              & \textcolor{red}{0.678}                          & 0.695              & \textcolor{red}{0.420}             & 0.949            & 0.083            & \textcolor{red}{0.802}             & 0.138                 & \textcolor{red}{0.648}   & 14s     \\ 
& ResNet18$^{++}$                        & 52                              & 0.689                          & 0.702              & 0.516             & 0.953            & 0.110            & 0.808             & 0.182       & 0.674 & 12s       \\ 
& ResNet50$^{++}$                        & 44                              & 0.738                          & 0.743              & 0.702             & 0.946            & 0.282            & 0.832             & 0.402          & 0.766  & 21s      \\ 
& VGG19$^{++}$                           & \textcolor{blue}{70}                              & \textcolor{blue}{0.839}                         & \textcolor{blue}{0.880}              & 0.746             & \textcolor{red}{0.886}            & \textcolor{blue}{0.736}            & \textcolor{blue}{0.883}             & \textcolor{blue}{0.741}           & \textcolor{blue}{0.913}    & 40s    \\ 
& ViT-B/16$^{++}$                      & \textcolor{blue}{70}                              & 0.811                          & 0.831              & \textcolor{blue}{0.750}             & 0.910            & 0.594            & 0.869             & 0.663          & 0.877   & 67s     \\ 
& EfficientNetV2-S$^{++}$             & 45                              & 0.768                          & 0.787              & 0.706             & 0.919            & 0.436            & 0.846             & 0.512        & 0.811 & 52s       \\ 
\bottomrule
\end{tabular*}
\footnotetext[1]{Refers to benign class.}
\footnotetext[2]{Refers to malignant class.}
\footnotetext[3]{Refers to average time per epoch, and is rounded up.}
\end{sidewaystable}

\begin{table}[h] 
\caption{Intermediate Results of Model$^{+}$ in Expt.\#2}\label{tab: Expt2_DR_results}
\begin{tabular}{@{}lccc@{}}
\toprule
Model	& Epoch & Val. Accuracy \footnotemark[1]{} & Avg. Time \footnotemark[2]{}\\
\midrule
DenseNet121$^{+}$ & 26  & 0.791 & 136s\\
DenseNet169$^{+}$	& 26	& 0.777 & 156s \\
InceptionV3$^{+}$	& 20	& 0.822 & 142s \\
MobileNet V2$^{+}$	& 17	& 0.808 & 120s \\
ResNet18$^{+}$	& 19 & 0.773 & 121s \\
ResNet50$^{+}$	& 38 & 0.743 & 106s \\
VGG19$^{+}$	& 32 & 0.738 & 150s \\
ViT-B/16$^{+}$	& 15 & 0.759 & 237s \\
EfficientNetV2-S$^{+}$	& 13 & 0.84 & 251s \\
\bottomrule
\end{tabular}
\footnotetext[1] {Validation accuracy.}
\footnotetext[2] {Average time per epoch, and is rounded up.}
\end{table}

Expt.\#1 investigate the effectiveness of one-step heterogeneous transfer learning. Though EfficientNetV2-S is the most efficient in time, converging in only 37 epochs, it performs worst in classification tasks with an overall accuracy of $82\%$. On the other hand, ViT-B/16 performs best in this scenario with an overall accuracy of $91.5\%$, as well as the best performed classifier from the ROC curves in Fig. \ref{subfig: ROC_Expt1}, although not as well in terms of efficiency. 

Expt.\#2 investigate the effectiveness of two-step heterogeneous transfer learning. The intermediate results are summarized in \autoref{tab: Expt2_DR_results}. 

VGG19 and ViT-B/16 both reach the maximum epoch limit, but VGG19 outperforms overall, achieving an accuracy of $83.9\%$ and emerging as the best classifier in Fig. \ref{subfig: ROC_Expt2}. It exhibits the highest recall for malignant class, indicating its superior ability to accurately identify malignant cases. However, it is important not to underestimate the precision rate of ViT-B/16 for the malignant class, as its high precision holds significant clinical value by reducing the likelihood of missing actual afflicted patients. Among the models, MobileNet V2 proves to be the most time-efficient; however, it performs the poorest in the laryngeal blood vessel classification task, with an accuracy of only $67.8\%$. 

\begin{figure*}[h]
\centering
\includegraphics[width=\linewidth]{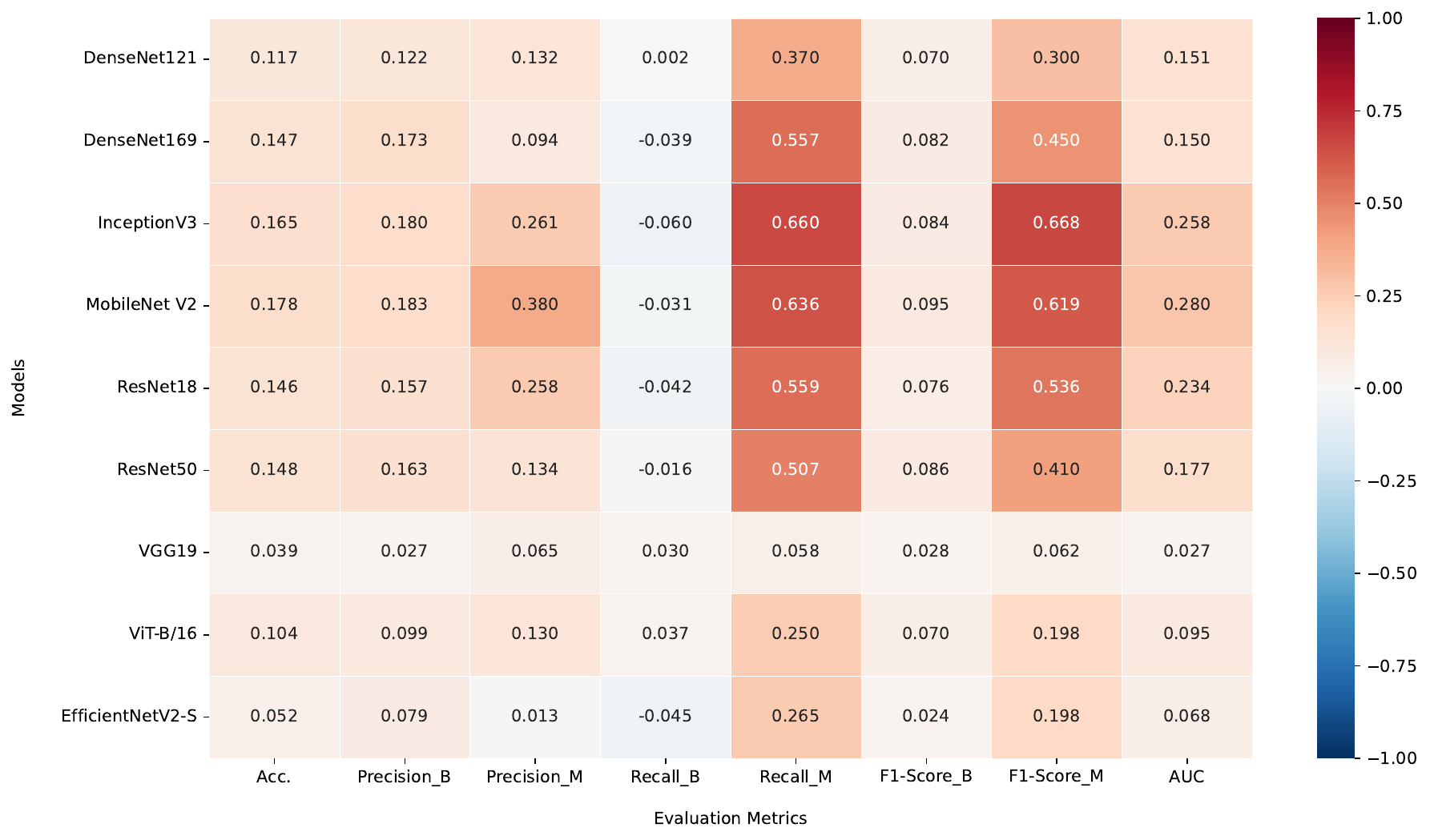}
\caption{Heatmap for the Results of Expt.\#1 Minus Results of Expt.\#2.} \label{fig: Difference_Exp1_Exp2}
\end{figure*}

Overall, there is a substantial drop in the performance of two-step heterogeneous transfer learning compared with one-step in this particular classification task. The average accuracy of nine models drops by $12\%$, and the average AUC drops around $16\%$. More details are presented as the heat map in Fig. \ref{fig: Difference_Exp1_Exp2}. The values in the heat map are the subtraction of the performance for Expt.\#1 and Expt.\#2. Therefore the specified range of each values is from $-1$ to $1$. It can be observed that the overall performance declined in Expt.\#2, with the malignant class contributing the most. The average percentage decrease in the malignant class is more significant than the average decrease in the benign class for each metric. Especially for recall and F1-Score, the malignant class shows an average of $31\%$ and $45\%$ more performance drop than the benign class. However, the benign class is slightly improved under the recall metric, around $2\%$. This figure illustrates that the same model pre-trained on ImageNet loses most of its capability to determine the malignant class after learning the vessel pattern from the intermediate domain. The possible reason is identify by the visualization of the model's attention and will be discussed in \autoref{sec: Discussion}.

\subsection{Results to Verify the Hypothesis (2)} \label{Results for H(2)}

The results of Expt.\#3 and Expt.\#4 are concluded in \autoref{tab: Expt3_and_Expt4_results}. And the ROC curves for these two experiments are presented in Fig. \ref{subfig: ROC_Expt3} and Fig. \ref{subfig: ROC_Expt4}. We select the run that is closest to the mean test accuracy presented in \autoref{tab: Expt3_and_Expt4_results} to generate the ROC curve.

\begin{sidewaystable}
\caption{Results of Expt.\#3. and Expt.\#4.} \label{tab: Expt3_and_Expt4_results}
\begin{tabular*}{\textwidth}{@{\extracolsep\fill}llcccccccccc}
\toprule%
&&&& \multicolumn{2}{c}{Precision} & \multicolumn{2}{c}{Recall} & \multicolumn{2}{c}{F1-Score} &\\
\cmidrule{5-10}%
Expt. & Model & Epoch & Accuracy & B\footnotemark[1]{} & M\footnotemark[2]{} & B & M & B & M & AUC & Avg. Time\footnotemark[3]{}\\
\midrule
\multirow{9}{*}{\#3}& DenseNet121$^{**}$                     & 32                              & 0.976                          & 0.983              & 0.961             & 0.982            & 0.964            & 0.983             & 0.962           & 0.998   & 32s   \\ 
& DenseNet169$^{**}$                     & 30                              & 0.969                          & 0.976              & 0.954             & 0.979            & 0.946            & 0.977             & 0.950       & 0.996  & 40s      \\ 
& InceptionV3$^{**}$                     & 27                              & 0.968                          & 0.974              & 0.954             & 0.979            & 0.943            & 0.977             & 0.948         & 0.997   & 34s     \\ 
& MobileNet V2$^{**}$                   & 32                              & 0.969                          & 0.979              & 0.947             & 0.976            & 0.954            & 0.977             & 0.950               & 0.997      & 16s  \\ 
& ResNet18$^{**}$                        & 28                              & 0.968                          & 0.975              & 0.951             & 0.978            & 0.945            & 0.977             & 0.948            & 0.995  & 13s      \\ 
& ResNet50$^{**}$                        & \textcolor{blue}{35}                              & 0.958                          & 0.969              & 0.935             & 0.970            & 0.931            & 0.970             & 0.933                & 0.993    & 26s    \\ 
& VGG19$^{**}$                           & 25                              & 0.975                          & 0.981              & 0.963             & \textcolor{blue}{0.983}            & 0.959            & 0.982             & 0.961         & 0.997   & 47s     \\ 
& ViT-B/16$^{**}$                      & 26                              & \textcolor{red}{0.951}                          & \textcolor{red}{0.964}              & \textcolor{red}{0.924}             & \textcolor{red}{0.966}            & \textcolor{red}{0.920}            & \textcolor{red}{0.965}             & \textcolor{red}{0.922}           & \textcolor{red}{0.983}  & 74s      \\ 
& EfficientNetV2-S$^{**}$             & \textcolor{red}{21}                              & \textcolor{blue}{0.980}                          & \textcolor{blue}{0.988}              & \textcolor{blue}{0.962}             & \textcolor{blue}{0.983}            & \textcolor{blue}{0.973}            & \textcolor{blue}{0.985}             & \textcolor{blue}{0.968}              & \textcolor{blue}{0.999}  & 57s      \\\midrule
\multirow{9}{*}{\#4}& DenseNet121\^                     & \textcolor{blue}{68}                                                  & 0.945                                              & 0.952                          & 0.928                          & 0.968                          & 0.894                          & 0.960                          & 0.910                        & 0.987   &30s                         \\ 
& DenseNet169\^                     & 58                                                  & 0.948                                              & 0.962                          & 0.917                          & 0.962                          & 0.917                          & 0.962                          & 0.917               & 0.987  & 34s                          \\ 
& InceptionV3\^                     & 48                                                  & \textcolor{blue}{0.964}                                              & \textcolor{blue}{0.976}                          & \textcolor{blue}{0.938}                          & 0.972                          & \textcolor{blue}{0.946}                          & \textcolor{blue}{0.974}                          & \textcolor{blue}{0.942}                        & \textcolor{blue}{0.993}   & 32s                         \\ 
& MobileNet V2\^                   & 42                                                  & 0.950                                              & 0.963                          & 0.920                          & 0.964                          & 0.920                          & 0.964                          & 0.920                 & 0.987   & 16s                         \\ 
& ResNet18\^                        & 51                                                  & 0.950                                              & 0.959                          & 0.929                          & 0.968                          & 0.910                          & 0.964                          & 0.919           & 0.988  & 12s                          \\ 
& ResNet50\^                        & 56                                                  & 0.947                                              & 0.956                          & 0.926                          & 0.967                          & 0.902                          & 0.961                          & 0.914            & 0.984     & 24s                       \\ 
& VGG19\^                           & 26                                                  & 0.936                                              & 0.948                          & 0.910                          & 0.960                          & 0.884                          & 0.954                          & 0.897           & 0.974   & 44s                         \\ 
& ViT-B/16\^                      & 37                                                  & 0.925                                              & 0.934                          & 0.902                          & \textcolor{red}{0.958}                          & 0.851                          & 0.946                          & 0.875         & 0.972          & 69s                 \\ 
& EfficientNetV2-S\^             & \textcolor{red}{21}                                                  & \textcolor{red}{0.695}                                              & \textcolor{red}{0.694}                          & \textcolor{red}{0.739}                          & \textcolor{blue}{0.994}                          & \textcolor{red}{0.035}                          & \textcolor{red}{0.817}                          & \textcolor{red}{0.068}                                      & \textcolor{red}{0.650}      & 58s                     \\
\bottomrule
\end{tabular*}
\footnotetext[1]{Refers to benign class.}
\footnotetext[2]{Refers to malignant class.}
\footnotetext[3]{Refers to average time per epoch, and is rounded up.}
\end{sidewaystable}

\begin{figure*}[h]
\centering
\subfloat[DenseNet121$^{**}$]{\includegraphics[width=0.33\linewidth]{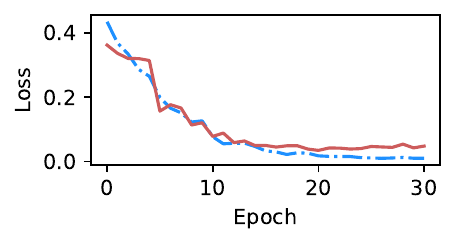}}
\hfil
\subfloat[DenseNet169$^{**}$]{\includegraphics[width=0.33\linewidth]{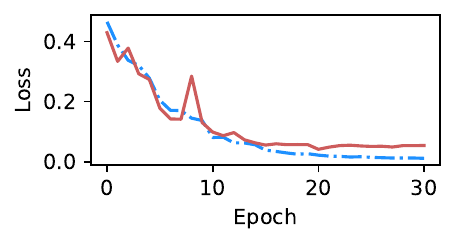}}
\hfil
\subfloat[InceptionV3$^{**}$]{\includegraphics[width=0.33\linewidth]{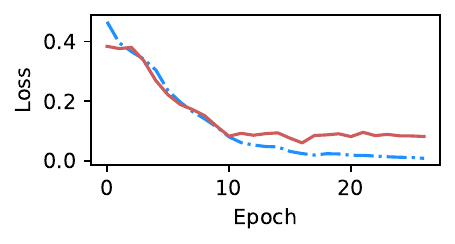}}
\\
\subfloat[MobileNet V2$^{**}$]{\includegraphics[width=0.33\linewidth]{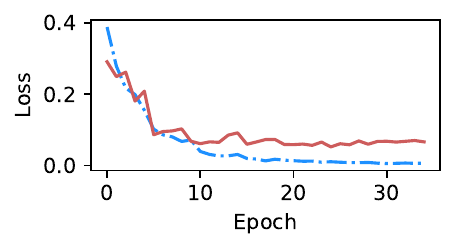}}
\hfil
\subfloat[ResNet18$^{**}$]{\includegraphics[width=0.33\linewidth]{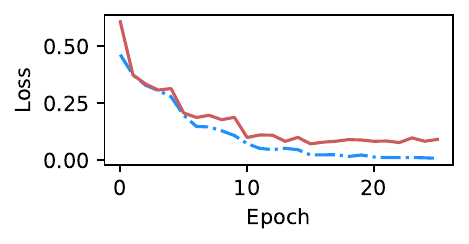}}
\hfil
\subfloat[ResNet50$^{**}$]{\includegraphics[width=0.33\linewidth]{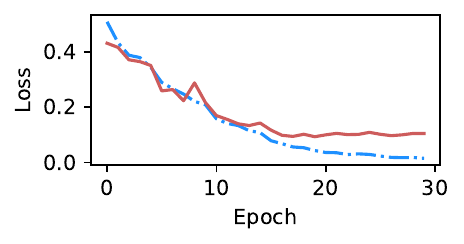}}
\\
\subfloat[VGG19$^{**}$]{\includegraphics[width=0.33\linewidth]{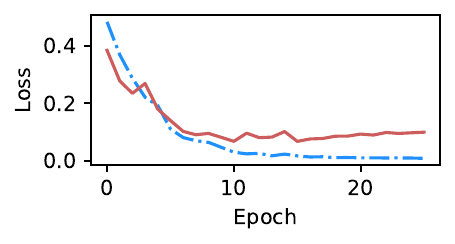}}
\hfil
\subfloat[ViT-B/16$^{**}$]{\includegraphics[width=0.33\linewidth]{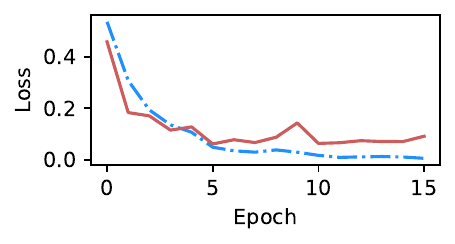}}
\hfil
\subfloat[EfficientNetV2-S$^{**}$]{\includegraphics[width=0.33\linewidth]{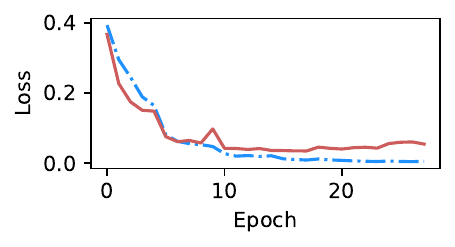}}
\\
\subfloat{\includegraphics[width=0.4\linewidth]{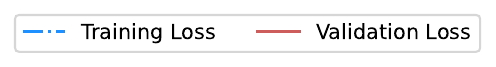}}
\caption{Training and Validation Loss Plots for Expt.\#3}
\label{fig: training and val loss for expt3}
\end{figure*}

In Expt.\#3, ImageNet is used as source domain. All models perform well, with accuracy above $95\%$ on the test set. Among them, the EfficientNetV2-S achieves the best performance ($98.0\%$ accuracy) with the least epochs (21). Meanwhile, the ViT-B/16 gets the most negligible result, with an accuracy of $95.1\%$ and AUC of $98.3\%$. VGG19 also performs well in terms of accuracy for benign class (recall $98.3\%$), and although it is ranked as the third place in terms of overall accuracy ($97.5\%$), the number of epochs of it is less than the second place model, the DenseNet121. The loss curves for Expt.\#3 are presented in Fig. \ref{fig: training and val loss for expt3}. Given that we conduct three runs for each experiment, we select the run that is closet to the average testing accuracy presented in the Expt.\#3 to generate the training and validation loss curves. 

In Expt.\#4, ImageNet is not used as source domain, therefore all the models are directly trained in the target domain. The most significant difference compared with the Expt.\#3 is the number of epochs. Expt.\#4 takes a much longer time to complete. The EfficientNetV2-S tells a different story to the Expt.\#3. In the Expt.\#4, the EfficientNetV2-S performs the worst (accuracy $69.5\%$, AUC $65.0\%$) while it completes within the least number of epochs (21). However, the EfficientNetV2-S performs the best for the recall of the benign class ($99.4\%$). The little discrimination for the malignant class is a fundamental reason for the low accuracy of this model, with a recall of only $3.5\%$ for the malignant class. The best performance for the models is the InceptionV3, with overall accuracy equals $96.4\%$ and AUC equals $99.3\%$. This result reverses the findings of the poorer performance in Expt.\#2. 

In conclusion, compared to training from scratch, the major advantage of using ImageNet as the source domain is a significant reduction in running time. Although the accuracy and AUC are slightly higher than training from scratch, the difference is not significant.

Generally speaking, all the models are more capable of identifying the benign class than the malignant class, regardless of the presence of an intermediate domain or whether they are trained from scratch. Mainly, ViT-B/16 shows excellent ability in one-step heterogeneous transfer learning. VGG19 is advanced in two-step heterogeneous transfer learning among the nine models. EfficientNetV2-S performs well when it requires well-prepared initial weights for its network and is thoroughly trained on the target domain. For training from scratch, InceptionV3 is more suitable in this scenario.

\section{Discussion}\label{sec: Discussion}
In this section, we use LayerCAM technique to delve into the possible reason for performance drop on the malignant class judgment of two-step heterogeneous transfer learning and introduce the Step-Wise Fine-Tuning technique to improve the performance.

\subsection{Analysis of Performance Drop using LayerCAM} \label{sec: LayerCAM}

For a long time, the actual inner operation of deep learning models has been a black box. However, with the help of the advent of attention mechanisms and the development of model visualization, it has allowed us to finally glimpse the inner mysteries of the models. The advent of LayerCAM permits us to visualize the class activation maps generated by different layers of the CNN, thus allowing us to observe the determination logic of the CNN for classification more nuancedly. LayerCAM employs gradient-based techniques to emphasize the significance of different locations within the feature map for a specific class. LayerCAM has the distinct ability to generate dependable class activation maps (locate the most relevant pixels to the target objects \citep{zhou2016learning}) from multiple layers of a CNN, rather than solely relying on the final convolutional layer. By adding these class-specific activation maps onto the original input image, LayerCAM produces a visual representation that effectively emphasizes the regions of the image that significantly contribute to the network's decision-making process. 

Given a predicted image, the class activation map calculated by the LayerCAM can be described using the following equations \citep{layerCAM}:
\begin{equation}
    M^c = ReLU (\sum_k (w_{ij}^{kc} \cdot A_{ij}^{k}))
\end{equation}
where,
\begin{align}\label{eq:6}
    w_{ij}^{kc} = ReLU(g_{ij}^{kc})
\end{align}

Specifically, $w_{ij}^{kc}$ represents the weight of the spatial coordinate $(i,j,)$ within the k-th feature map with c channels. The gradients of this coordinate $g_{ij}^{kc}$ remain if it is greater than zero; otherwise, it turns to zero. Then, LayerCAM multiplies the activation value of that coordinate with the weight in the Equation (\ref{eq:6}) and linearly combines all channel dimensions to calculate the final class activation map $M^c$. 

In this work, we use the commonly used deep learning model in the image classification field, say the ResNet18, to perform the task. ResNet18 contains four extensive basic modules called Layer One to Four, in Pytorch. By combining the LayerCAM with the ResNet18, the attention area of the model in each Layer can be represented in the image. Model pays more attention to areas that are highlighted in red. The best-performing ResNet18 among the previous four experimental scenarios is selected as the benchmark, the ResNet18$^{**}$ in Expt.\#3. Additionally, ResNet18$^{++}$ from Expt.\#2 shows the most significant performance drop in classification, therefore it is compared with the benchmark ResNet18$^{**}$ to investigate the possible reason for the performance degradation.

\begin{figure*}[h]
\centering
\subfloat[Layer 1]{\includegraphics[width=0.25\linewidth]{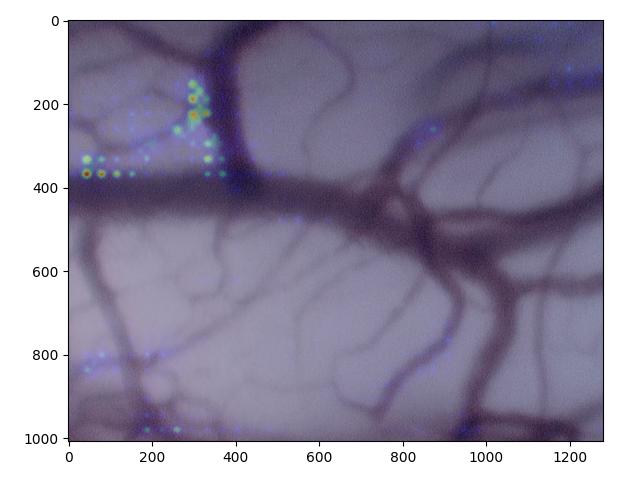}}
\hfil
\subfloat[Layer 2]{\includegraphics[width=0.25\linewidth]{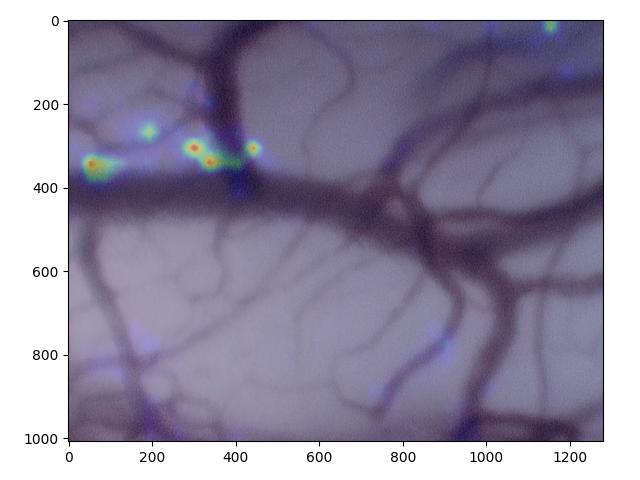}}
\hfil
\subfloat[Layer 3]{\includegraphics[width=0.25\linewidth]{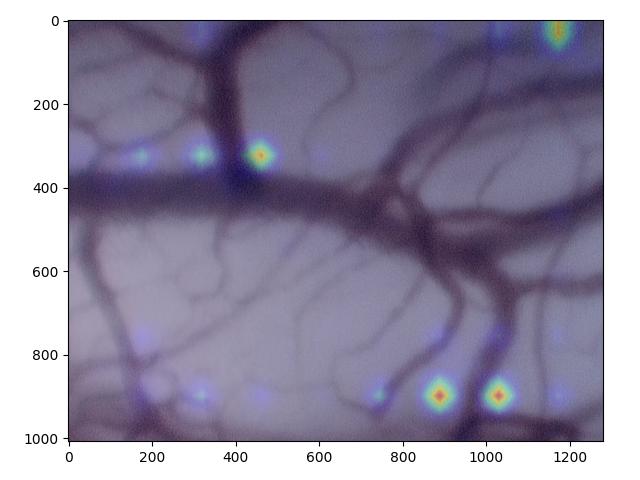}}
\hfil
\subfloat[Layer 4]{\includegraphics[width=0.25\linewidth]{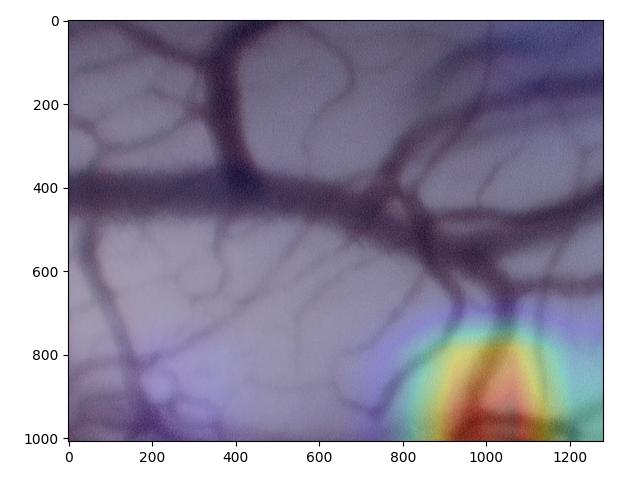}}
\caption{Visualize the Area of Interest of ResNet18$^{**}$ on Benign Example using layerCAM.}
\label{fig: resnet18**_benign}
\end{figure*}

\begin{figure*}[h]
\centering
\subfloat[Layer 1]{\includegraphics[width=0.25\columnwidth]{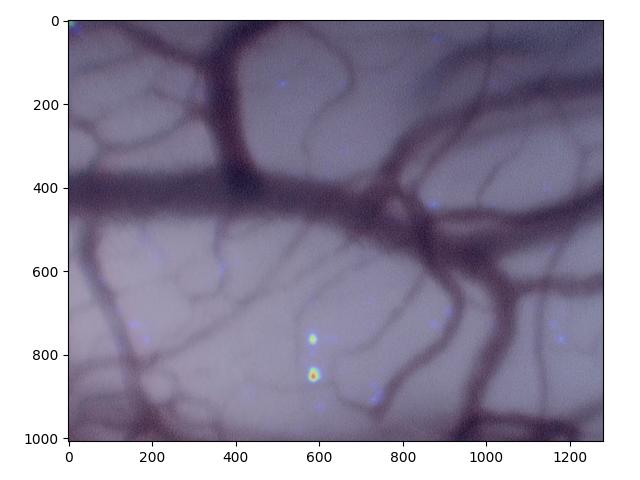}}
\hfil
\subfloat[Layer 2]{\includegraphics[width=0.25\columnwidth]{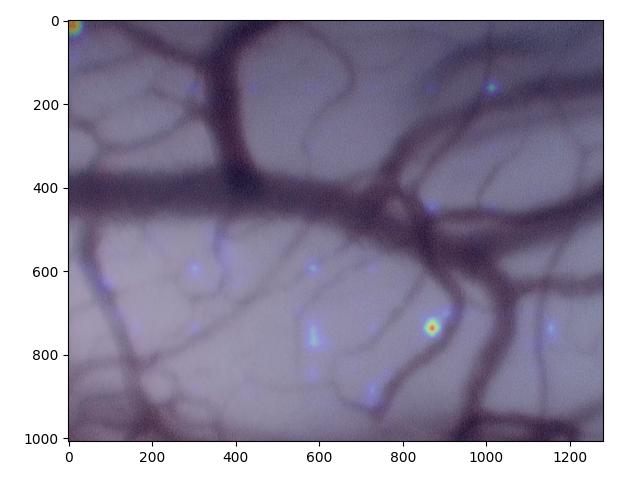}}
\hfil
\subfloat[Layer 3]{\includegraphics[width=0.25\columnwidth]{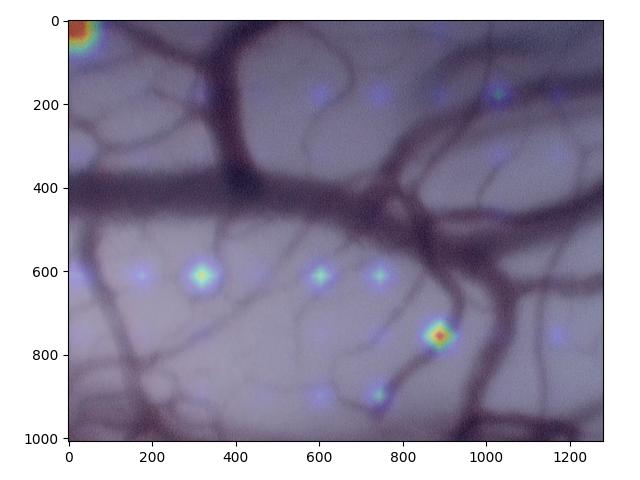}}
\hfil
\subfloat[Layer 4]{\includegraphics[width=0.25\columnwidth]{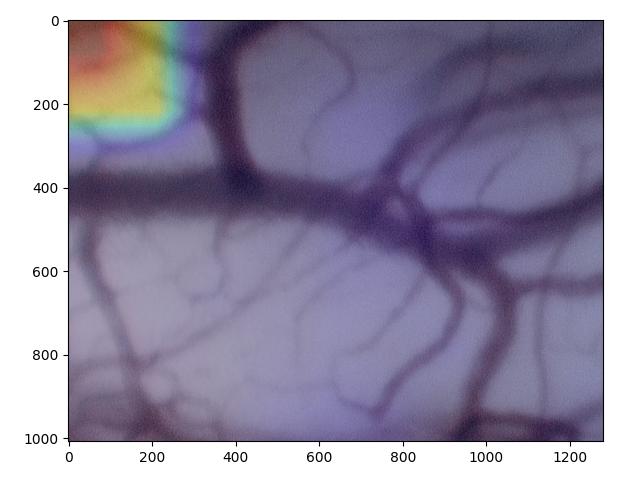}}
\caption{Visualize the Area of Interest of ResNet18$^{++}$ on Benign Example using layerCAM.}
\label{fig: resnet18++_benign}
\end{figure*}

We randomly select a benign-labeled image (image number Patient002\_P002 (107)) from the test set. Then the LayerCAM is used to visualize the benign class based on ResNet18$^{**}$ and ResNet18$^{++}$, shown in Fig. \ref{fig: resnet18**_benign} and Fig. \ref{fig: resnet18++_benign}. Both ResNet18$^{**}$ and ResNet18$^{++}$ deliver a correct prediction on this image. It can be observed that ResNet18$^{**}$ starts focusing more on the contours of the major vessels and gradually moving the focus to the lower right corner of the image as the number of layers increases, with the thicker vessels being the primary basis for judgment. ResNet18$^{++}$ applies the acquired vascular knowledge to the classification prediction task of the target domain by learning the intermediate domain, it puts more effort into capturing the fine vessels used as a basis for judgment. It can be inferred that after learning in the intermediate domain, the ResNet18 holds the ability to locate the fine vessels. 

\begin{figure*}[h]
\centering
\subfloat[Layer 1]{\includegraphics[width=0.25\columnwidth]{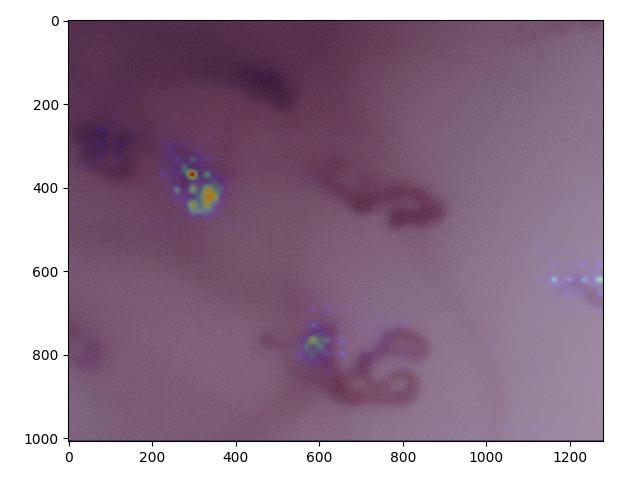}\label{subfig: Resnet18**_layer1_Malignant}}
\hfil
\subfloat[Layer 2]{\includegraphics[width=0.25\columnwidth]{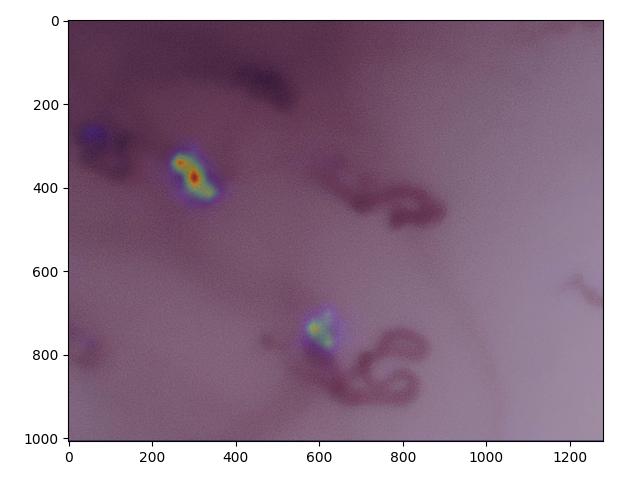}}
\hfil
\subfloat[Layer 3]{\includegraphics[width=0.25\columnwidth]{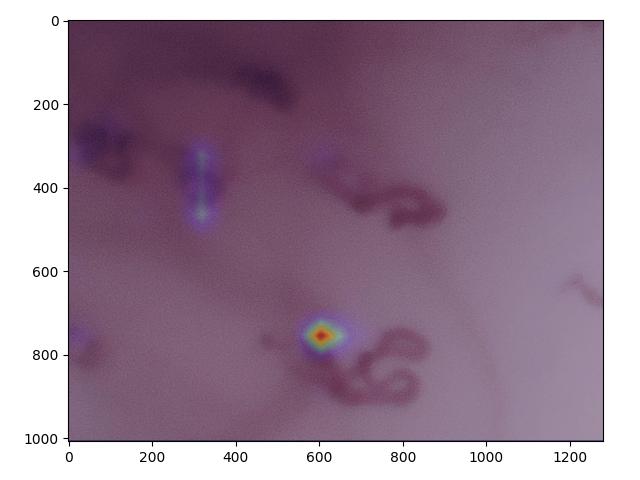}}
\hfil
\subfloat[Layer 4]{\includegraphics[width=0.25\columnwidth]{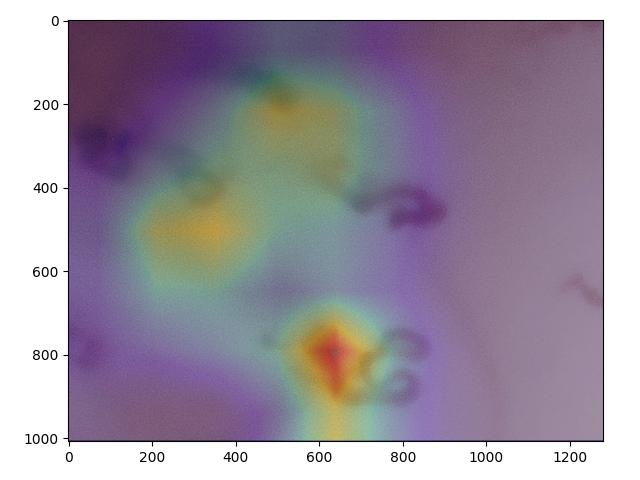}}
\caption{Visualize the Area of Interest of ResNet18$^{**}$ on Malignant Example using layerCAM.}
\label{fig: Resnet18**_malignant}
\end{figure*}

\begin{figure*}[h]
\centering
\subfloat[Layer 1]{\includegraphics[width=0.25\columnwidth]{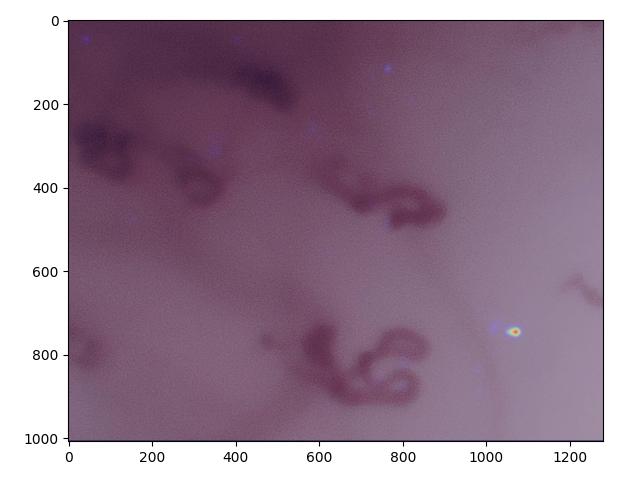}}
\hfil
\subfloat[Layer 2]{\includegraphics[width=0.25\columnwidth]{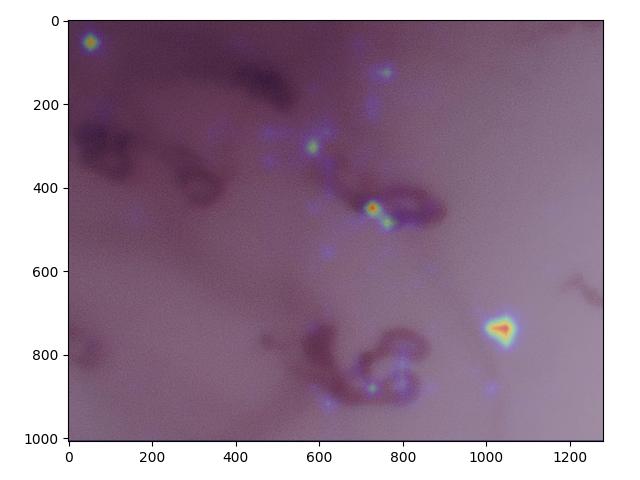}\label{subfig: Resnet18++_layer2_Malignant}}
\hfil
\subfloat[Layer 3]{\includegraphics[width=0.25\columnwidth]{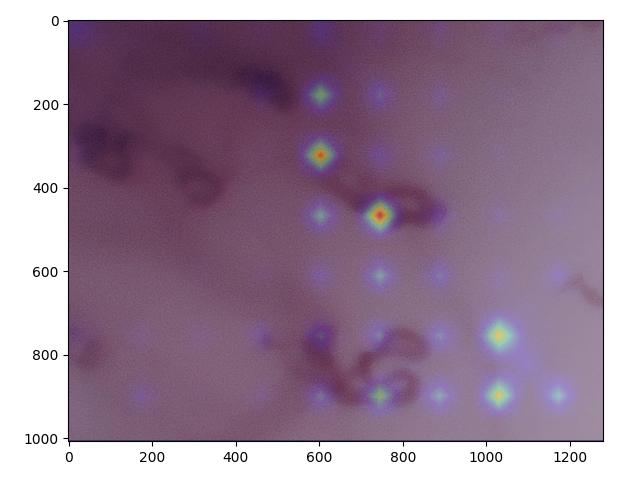}\label{subfig: Resnet18++_layer3_Malignant}}
\hfil
\subfloat[Layer 4]{\includegraphics[width=0.25\columnwidth]{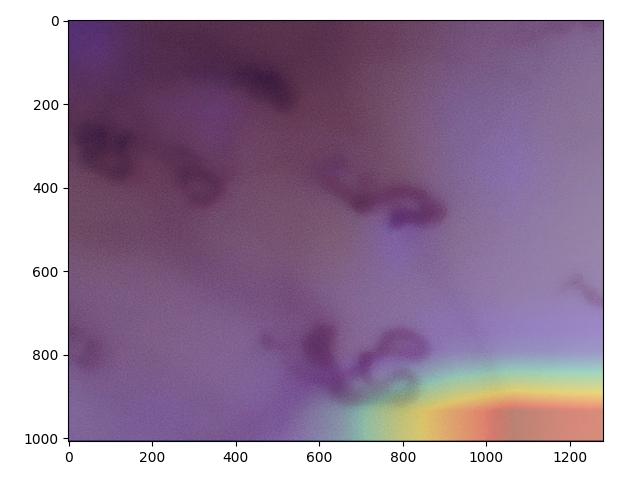}}
\caption{Visualize the Area of Interest of ResNet18$^{++}$ on Malignant Example using layerCAM.}
\label{fig: Resnet18++_malignant}
\end{figure*}

A malignant-labeled image is randomly selected from the test set, with image number Patient013\_P013 (36). The LayerCAM visualization of the malignant class based on ResNet18$^{**}$ and ResNet18$^{++}$ are represented in Fig. \ref{fig: Resnet18**_malignant} and Fig. \ref{fig: Resnet18++_malignant}. Unlike the benign classification, this time, ResNet18$^{**}$ gives a correct prediction for the image while ResNet18$^{++}$ does not. As can be seen in Fig. \ref{subfig: Resnet18**_layer1_Malignant}, ResNet18$^{**}$ puts its attention on the edges of the twisted blood vessels at the beginning. As the number of layers increases, the model concentrates more on the twisted or tadpole-shaped blood vessels which are highlighted by LayerCAM. However, for ResNet18$^{++}$, the attention of the model starts from Layer1 to Layer3, showing an increasing trend of dispersion and a grid alike appearance. Contrary to the trend of increasing concentration in ResNet18$^{**}$. Then finally, ResNet18$^{++}$ focuses on the blank space at the lower right corner of the image. The comparison of Fig. \ref{fig: Resnet18**_malignant} and Fig. \ref{fig: Resnet18++_malignant} suggests that, following learning in the intermediate domain, ResNet18 does not seize the knowledge to distinguish twisted, entangled blood vessels, therefore the classification performance on this class drops.

To sum up, images from the target dataset do not always follow a radial pattern, and some images show a large angle bend for blood vessels as the vascular pattern analysis by \citet{esmaeili2019novel}. Through the analysis of LayerCAM visualization outcomes, we speculate that in two-step heterogeneous transfer learning, ResNet18 loses the ability to discriminate twisted and tangled vessels while gaining the ability to capture fine vessels. This speculation applies to generalize to other models that have significant performance drops for predicting the malignant class in Expt.\#2.

\subsection{Step-Wise Fine-Tuning} \label{sec: SWFT}

\begin{figure*}[t]
\centering
\subfloat[Bar-Line Combination Chart of Step-Wise Fine-Tuning for ResNet18$^{++}$]{\includegraphics[width=0.45\linewidth]{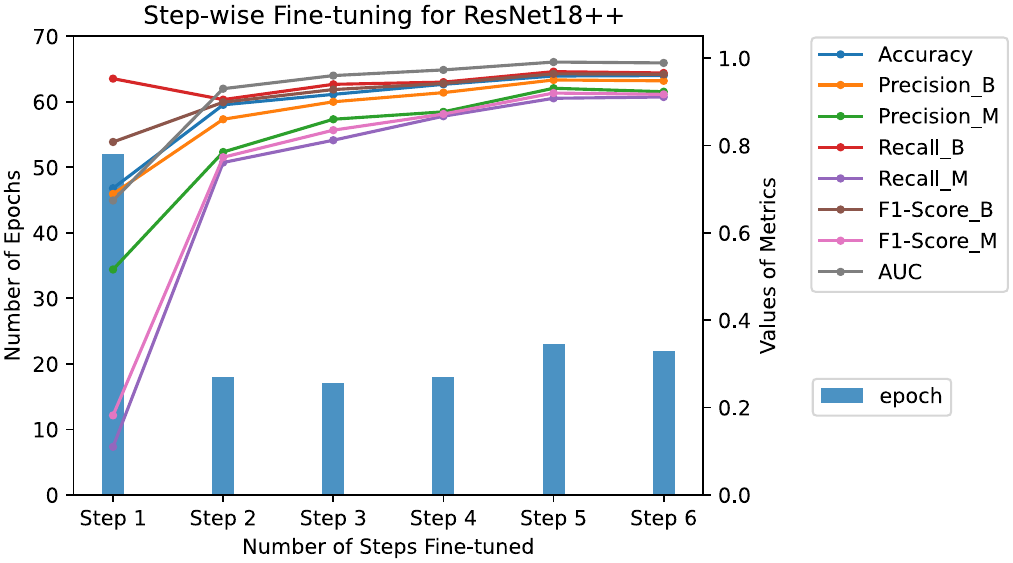}\label{subfig: ResNet18 SWFT}}
\hfil
\subfloat[Bar-Line Combination Chart of Step-Wise Fine-Tuning for ResNet50$^{++}$]{\includegraphics[width=0.45\linewidth]{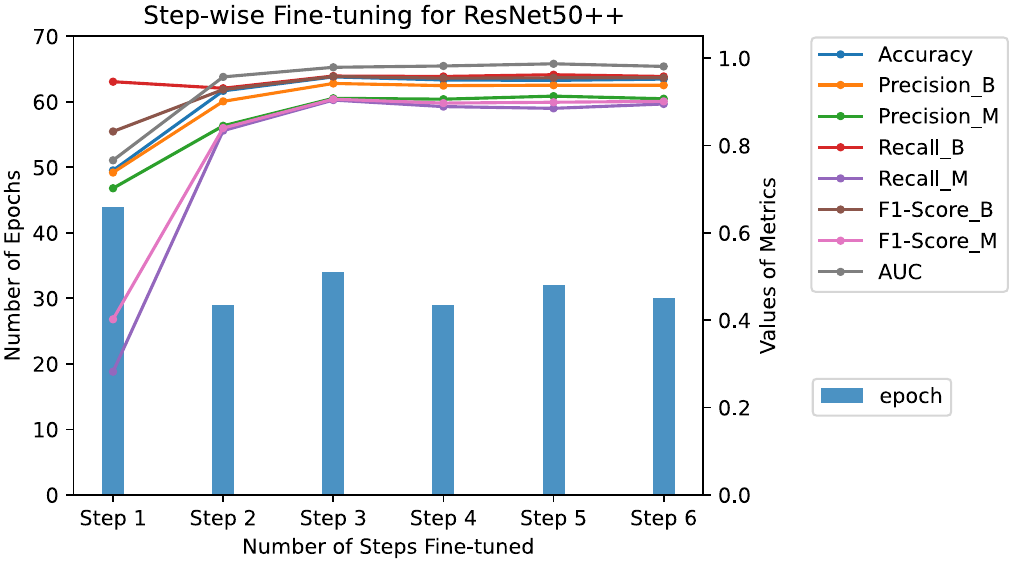}\label{subfig: ResNet50 SWFT}}
\caption{Bar-Line Combination Chart of Step-Wise Fine-Tuning for ResNet.}
\label{fig: Bar-Line ResNet SWFT}
\end{figure*}

\begin{table}[h]
\caption{Results of Step-Wise Fine-Tuning for ResNet18$^{++}$} \label{tab: SWFT_ResNet18_Result}
\begin{tabular*}{\textwidth}{@{\extracolsep\fill}cccccccccccc}
\toprule%
&&& \multicolumn{2}{c}{Precision} & \multicolumn{2}{c}{Recall} & \multicolumn{2}{c}{F1-Score} &\\
\cmidrule{4-9}%
Step & Epoch & Accuracy & B\footnotemark[1]{} & M\footnotemark[2]{} & B & M & B & M & AUC & Avg. Time\footnotemark[3]{}\\
\midrule
1 & 52 & 0.689 & 0.702 & 0.516 & 0.953 & 0.11 & 0.808 & 0.182 & 0.674 & 12s \\
2 & 18 & 0.86 & 0.893 & 0.785 & 0.905 & 0.761 & 0.899 & 0.773 & 0.93 & 14s\\
3 & 17 & 0.9 & 0.917 & 0.86 & 0.94 & 0.812 & 0.928 & 0.835 & 0.96 & 15s\\
4 & 18 & 0.921 & 0.94 & 0.877 & 0.945 & 0.867 & 0.942 & 0.872 & 0.973 & 15s\\
5 & 23 & 0.95 & 0.959 & 0.931 & 0.969 & 0.908 & 0.964 & 0.92 & 0.991 & 14s\\
6 & 22 & 0.948 & 0.96 & 0.923 & 0.966 & 0.911 & 0.963 & 0.917 & 0.989 & 13s \\
\bottomrule
\end{tabular*}
\footnotetext[1]{Refers to the benign class.}
\footnotetext[2]{Refers to the malignant class.}
\footnotetext[3]{Refers to average time per epoch, and is rounded up.}
\end{table}

\begin{table}[h]
\caption{Results of Step-Wise Fine-Tuning for ResNet50$^{++}$} \label{tab: SWFT_ResNet50_Result}
\begin{tabular*}{\textwidth}{@{\extracolsep\fill}cccccccccccc}
\toprule%
&&& \multicolumn{2}{c}{Precision} & \multicolumn{2}{c}{Recall} & \multicolumn{2}{c}{F1-Score} &\\
\cmidrule{4-9}%
Step & Epoch & Accuracy & B\footnotemark[1]{} & M\footnotemark[2]{} & B & M & B & M & AUC & Avg. Time \footnotemark[3]{}\\
\midrule
1 & 44 & 0.738 & 0.743 & 0.702 & 0.946 & 0.282 & 0.832 & 0.402 & 0.766 & 21s \\
2 & 29 & 0.901 & 0.925 & 0.845 & 0.931 & 0.834 & 0.928 & 0.84 & 0.957 & 27s\\
3 & 34 & 0.942 & 0.957 & 0.908 & 0.959 & 0.904 & 0.958 & 0.906 & 0.979 & 27s\\
4 & 29 & 0.937 & 0.95 & 0.906 & 0.958 & 0.889 & 0.954 & 0.897 & 0.982 & 37s\\
5 & 32 & 0.938 & 0.949 & 0.913 & 0.962 & 0.885 & 0.955 & 0.899 & 0.987 & 26s\\
6 & 30 & 0.938 & 0.952 & 0.907 & 0.958 & 0.895 & 0.955 & 0.901 & 0.981 & 25s \\
\bottomrule
\end{tabular*}
\footnotetext[1]{Refers to the benign class.}
\footnotetext[2]{Refers to the malignant class.}
\footnotetext[3]{Refers to average time per epoch, and is rounded up.}
\end{table}

This subsection introduces Step-Wise Fine-Tuning technique for improving the unsatisfactory performance of two-step heterogeneous transfer learning in Expt.\#2. 

Since we use LayerCAM to visualize the attention areas for ResNet18 in \autoref{sec: LayerCAM}, to better cooperate with the LayerCAM visualization and ResNet's modules in Pytorch, we propose a fine-tuning technique called Step-Wise Fine-Tuning.

Following this Step-Wise Fine-Tuning design, we conduct experiments with ResNet18$^{++}$ and ResNet50$^{++}$ on the target domain (CE-NBI dataset). The metrics from Expt.\#2 are extended here, and similarly, the average value of three runs is recorded for each metric, then presented in the figures below. The blue bar represents the number of epochs, and the various colored lines represent the metrics. The result of ResNet18$^{++}$ and ResNet50$^{++}$ is shown in Fig. \ref{subfig: ResNet18 SWFT} and Fig. \ref{subfig: ResNet50 SWFT}, respectively. B and M in the legend represent the benign and malignant classes. We also summarize the Step-Wise Fine-Tuning results of ResNet18$^{++}$ and ResNet50$^{++}$ in \autoref{tab: SWFT_ResNet18_Result} an \autoref{tab: SWFT_ResNet50_Result}.

It can be observed from Fig. \ref{fig: Bar-Line ResNet SWFT} that the performance of the classification task improves significantly with the increase in the number of steps, especially from Step 1 to Step 2. From the metrics perspective, only fine-tuning the fully connected layer is less efficient. It takes the longest time to complete the task and performs least unsatisfied. 

Specifically, our proposed SWFT also incorporates the visualization of the LayerCAM as a reference for the number of steps that should be involved in SWFT. As illustrated in Fig. \ref{fig: Resnet18++_malignant}, after the intermediate domain, ResNet18$^{++}$ focuses on empty regions rather than the blood vessels (as depicted in Fig. \ref{subfig: Resnet18**_layer1_Malignant}). Also, the attention areas in Fig. \ref{subfig: Resnet18++_layer2_Malignant} and Fig. \ref{subfig: Resnet18++_layer3_Malignant} tend to be grid-like. This observation suggests that SWFT should be fine-tuned to step 5. The efficacy of this approach is substantiated by \autoref{tab: SWFT_ResNet18_Result}, which demonstrates that the performance achieved at step 5 surpasses not only the performance of step 1 (increases by 26.1\%) but even that of fine-tuning the entire network, both in accuracy and AUC. Still, this performance is 1.8\% less accurate compared to Expt.\#3, but it is considerably closer. 

For ResNet50$^{++}$, fine-tuning the model up to Step 3 seems reasonable enough since it can be obtained from the \autoref{tab: SWFT_ResNet50_Result}, Step 3 has the highest accuracy and comparable AUC while relatively short running time. Based on the result of Step 3 and Step 1 (Expt.\#2), accuracy improves by 20.4\% with a 1.9\% reduction in running time. Still, this performance is 1.6\% less accurate compared to Expt.\#3, but it is close. A conclusion can be drawn that a shallower deep learning model needs more steps to fine-tune than a deeper model.

\section{Conclusion}\label{sec: Conclusion}

In this work,  to explore the performance of one-
step and two-step heterogeneous transfer learning, we propose and validate the effectiveness of using color fundus photographs of the diabetic retina dataset as an intermediate domain for two-step heterogeneous learning in classifying laryngeal
vascular images using nine deep-learning models. We use ImageNet as the source domain, the DR dataset as the intermediate domain, and CE-NBI dataset as the target domain. 

The experiments demonstrate that models diminish the ability to distinguish the malignant class after learning in the intermediate domain, reflected in the evaluation metrics, such as precision, recall, AUC, and accuracy. The answer of the hypothesis (1) is negative. With the help of LayerCAM, the area of interest of the model is visualized layer by layer, finding that in the two-step heterogeneous transfer learning, the model is struggling to pay attention to the twisted, tangled vessels, which are the essential character for distinguishing the malignant class. The possible reason for the unsatisfactory performance of two-step heterogeneous transfer learning is the different features of the blood vessels. Most of the vessels in the intermediate domains show a radial pattern, resulting in the model not learning a pattern of similar twisted and tangled vessels. Therefore, in the two-step heterogeneous transfer learning of blood vessels, the features of the lesions in the intermediate domain and the target domain should be similar for the learning to be effective. Furthermore, we propose a Step-Wise Fine-Tuning technique to improve the performance of two-step heterogeneous transfer learning in the second step. By stacking forward the number of layers involved in fine-tuning, the performance of the model is greatly improved. Meanwhile, we also explore the answer for the hypothesis (2). The experiment results state that the performance of using ImageNet as the source domain is slightly better than training from scratch, but the improvement is not significant. 

Despite our efforts, some limitations still exist within our work. One of the biases affecting the results could be caused by the fact that the features of the blood vessels in the intermediate domain are skewed towards the benign class from the target domain, lacking the features used to identify the malignant class from the target domain. Hence, in two-step heterogeneous transfer learning, the intermediate domain should contain more features to cover the features needed in the target domain. Our work concludes that the blood vessels in the intermediate domain should contain features of radial, twisted vascular patterns to serve the target domain better. This conclusion should also be followed in the subsequent selection of the intermediate domain in further laryngeal blood vessel classification studies. 

Another possible bias that might affect the results could be the limited dataset available. Since our work focuses on supervised learning, compared to unlabeled data or using pseudo-labeled data, supervised learning enables the model to learn more accurate knowledge. However, at the same time, it leads to a greater scarcity of data. Therefore, in the future, it is worth exploring the use of self-supervised learning in two-step heterogeneous transfer learning to increase the amount of data used for training, not only restricted by the blood vessels but also to investigate more possibilities.

\bmhead{Acknowledgements}

We would like to express our gratitude to Doctor Zhang Binghuang, the attending physician of the Department of Otolaryngology Head and Neck Surgery from the First Affiliated Hospital of Xiamen University, for his professional advice.

\section*{Declarations}

The authors declare that they have no known competing financial interests or personal relationships that could have appeared to influence the work reported in this paper.

\begin{itemize}
\item Funding: This work was supported by Macao Polytechnic University under grant number RP/FCA-04/2022.
\item Conflict of interest/Competing interests (check journal-specific guidelines for which heading to use): Not applicable.
\item Ethical and informed consent for data used: Not applicable.
\item Data availability: CE-NBI dataset was deposited into Zenodo under doi number 10.5281/zenodo.6674034 and is available at the following URL: https://zenodo.org/records/6674034. Example from: https://www.mdpi.com/1424-8220/21/23/8157. Diabetic Retinopathy Detection was deposited into Kaggle and is available at the following URL: https://www.kaggle.com/competitions/diabetic-retinopathy-detection/overview. 
\item Author contribution: Conceptualization: Xinyi Fang; Methodology: Xinyi Fang; Formal analysis and investigation: Xinyi Fang, Chak Fong Chong, Kei Long Wong; Writing - original draft preparation: Xinyi Fang; Writing - review and editing: Xinyi Fang, Xu Yang, Chak Fong Chong, Kei Long Wong, Yapeng Wang, Tiankui Zhang, Sio-Kei Im; Funding acquisition: Xu Yang; Resources: Yapeng Wang, Sio-Kei Im, TianKui Zhang; Supervision: Yapeng Wang.
\end{itemize}

\bibliography{sn-bibliography}

\end{document}